\documentclass[letterpaper]{article}
\usepackage{aaai2027}
\usepackage[hyphens]{url}
\usepackage{graphicx}
\usepackage{natbib}
\usepackage{caption}
\usepackage{algorithm}
\usepackage{algorithmic}
\usepackage{booktabs}
\usepackage{amsmath,amssymb}
\usepackage{subcaption}
\usepackage{booktabs}
\usepackage{multirow}
\usepackage{tabularx}
\usepackage{listings}
\usepackage{float}
\usepackage[breaklinks,colorlinks,citecolor=blue,linkcolor=blue,urlcolor=blue]{hyperref}

\definecolor{promptbackground}{RGB}{247,247,247}
\definecolor{promptframe}{RGB}{210,210,210}
\lstdefinestyle{prompttemplate}{
  breakautoindent=true,
  breakindent=2em
}

\providecommand{\Description}[1]{}
\title{HarnessNet: Feedback-Driven Hierarchical \\ Joint Harness Evolution for Multi-Agent Systems}

\author{Zhaoyuan Qiu \quad Sachith Seneviratne \quad Yu Xia \quad Saman Halgamuge\\
\vspace{1\baselineskip}
\small University of Melbourne, Melbourne, Australia\\
\small \texttt{zhaoyuanqiu@student.unimelb.edu.au}}

\affiliations{}
\nocopyright
\begin{document}
\maketitle
\begin{abstract}

AI agents can evolve with feedback from task environments to revise their harnesses, the softwares that manage model inputs, tool use, memory, and execution, while keeping the underlying foundation models frozen. Recent studies focus on this process for single agents, leaving underexplored the harness evolution in Multi-agent System (MAS), which requires coordinating harness changes across multiple agents. However, using a centralized global optimizer, the straightforward extension from single agent harness evolution, faces a growing context and workload as the MAS scale increases. Access restrictions introduce further challenges: agents may be unable to share internal information with a common optimizer, and some agents may lack permission to access the environment feedback which is for the entire MAS. To address these challenges, we propose HarnessNet, a framework for hierarchical harness evolution in MAS. Its basic unit, Agentron, consists of an evolvable forward harness, a foundation model, and a local optimizer that can observe and modify only that forward harness. Multiple Agentrons are orchestrated into a layered graph, and feedback is layerwisely propagated backward from the task environment so that each optimizer only receives node-specific evolution signals from Agentrons in adjacent downstream layer. Joint harness evolution therefore be divided into smaller, coordinated tasks with separate contexts. Each optimizer needs not access trajectories of all Agentrons or raw environment feedback. This reduces the context and workload handled per optimizer call, and keeps optimizer access local. We conduct experiments on AgentBench with multiple foundation models to demonstrate the feasibility of training Agentron and minimal HarnessNets. A comparison study against single global optimizer approach shows that HarnessNet achieves comparable performance gains while limiting optimizer access, with fewer tokens and less time required per optimizer call. A migration study further shows that the evolved harnesses remain effective after replacing the foundation models, suggesting good transferability.

\end{abstract}

\section{Introduction}
The behavior of Large Language Model (LLM) based agents depends on both the underlying models and the software infrastructure~\cite{hu2025adas,livesweagent2025}. Agent development therefore places growing attention on refining the software that organizes and directs model capabilities. This trend reflects an emerging paradigm shift in AI system optimization: from a primary focus on model weights toward model-external components that govern system behavior, release or extend the capabilities attainable with a given foundation model. Examples include prompt engineering~\cite{liu2023pretrain,schulhoff2024promptreport}, context engineering~\cite{mei2025context,anthropic2025context}, and, more recently, harness engineering~\cite{young2025harness,lopopolo2026harness}. Expressed in natural language and code, these software components and their revisions are directly accessible to human inspection and audit, supporting efforts to improve system security and controllability. 
Viewing the harness itself as an optimization target motivates \textbf{harness evolution}, which improves the agentic system through iterative harness refinement while freezing the foundation models.

Recently harness evolution studies focus on the  single-agent scenario~\cite{metaharness2026,lifeharness2026,ahe2026,continualharness2026,adaptiveautoharness2026,jiang2026harnessevolve}, leaving the \textbf{joint harness evolution in Multi-agent System (MAS)} underexplored. In single-agent approaches, people use a coding agent as the optimizer. This optimizer analyzes the agent’s execution trajectories and modifies its harness based on environment feedback. The natural extension to MAS is to have a single centralized optimizer which observes the trajectories of all agents and modifies their harnesses together. However, such a \textbf{global optimizer} extension faces challenges: \textbf{(1)} as the system scales up, this one-for-all design has a growing context length and workload burden, which can degrade its effect~\cite{du2025context,lindenbauer2025complexity,hao2026evolveteamcollaborativeselfevolution};  \textbf{(2)} in scenario with distinct access constraints, agents may not be able to expose their internal trajectories to a shared optimizer~\cite{bagdasarian2026agenticprivacy,juneja2025magpie,elyagoubi2026agentleak}, or directly access the environment feedback of the whole system. These limitations motivate a solution that decomposes MAS harness evolution into small, semi-independent local tasks while coordinating harness updates across agents, using only local feedback. In each task, optimizer visibility and authority are restricted to the agent being updated and thus respect access boundaries.

To address the gap and challenges, we propose HarnessNet, a framework of hierarchical, feedback-driven joint harness evolution for MAS. Its fundamental agentic unit, Agentron, couples an evolvable agent harness with a local optimizer instance, whose access and authority are confined to that unit. HarnessNet organizes Agentrons into a layered supergraph, and introduces hierarchical, access-constrained joint evolution of agent harnesses. Downstream task feedback is progressively translated into node-specific optimization signals, enabling coordinated updates without centralizing all execution traces. Experiments on AgentBench~\cite{liu2024agentbench} demonstrate significant improvements in task performance through iterative hierarchical harness evolution. Further migration studies show that evolved harnesses retain useful task-specific mechanisms across different foundation models. Contributions include:
\begin{enumerate}
    \item We introduce HarnessNet for feedback-driven joint harness evolution in MAS, and its basic unit Agentron. 

    \item We develop an automated task-specific HarnessNet initialization procedure and train several minimal HarnessNets on AgentBench tasks. Results show that automated initialization is feasible, so is MAS-scale harness evolution through hierarchical feedback-driven process.

    \item We assess the transferability of evolved HarnessNet systems through migration studies across foundation models. Results indicate that task-specific rules learned within harnesses can remain jointly effective on different models.

    \item We compare approaches with hierarchical evolution and centralized global optimizer. Hierarchical evolution produces competitive training effects while constraining local optimizer access and reducing per-optimizer-call cost.
\end{enumerate}

\section{Related Work}

\subsection{Agent Improvement Beyond Model Weights}
Beyond the foundation model, an agent relies on semantic artifacts that shape its behavior and execution strategy. Refining these artifacts offers a way to improve agent performance without updating model weights. ProTeGi~\cite{protegi2023}, OPRO~\cite{opro2024}, PromptAgent~\cite{promptagent2024}, and GEPA~\cite{gepa2025} optimize task instructions through iterative search and feedback, while MemAPO~\cite{memapo2026} accumulates reusable strategies and failure patterns to guide prompt refinement. Context-oriented approaches include Context Language Models~\cite{shao2026contextlanguagemodels} and Agentic Context Engineering~\cite{ace2025}, complemented by ACON's feedback-driven context compression~\cite{kang2026acon} and HiAgent's hierarchical working memory management~\cite{hiagent2025}. Reflexion~\cite{reflexion2023}, A-MEM~\cite{amem2025}, and Dynamic Cheatsheet~\cite{dynamiccheatsheet2025} improve the reuse of past experience through textual reflection, memory organization, and persistent strategy accumulation. AutoSkill~\cite{autoskill2026}, MUSE-Autoskill~\cite{museautoskill2026}, CoEvoSkills~\cite{coevoskills2026}, SkillGrad~\cite{skillgrad2026}, and MetaSkill-Evolve~\cite{metaskillevolve2026} further explore the refinement of reusable agent skills.  These methods establish prompts, context, and skills as targets for agent improvement beyond model weights. Moreover, TextGrad~\cite{textgrad2024} backpropagates textual feedback instead of gradients to optimize compound AI systems, introducing human-friendly evolution signals.

\subsection{Agent Harness Evolution}
Agent harness evolution extends individual artifact refinement to the agent harness, the broader agent software infrastructure, as the optimization target. SICA~\cite{selfimprovingcoding2025}, Darwin G{\"o}del Machine~\cite{dgm2025}, and MOSS~\cite{moss2026} show that coding agents can revise their own software to create agent substrates. Meta-Harness searches over complete harness implementations using source code,
traces, and scores~\cite{metaharness2026}. AHE structures harness components
and execution evidence to support observable edits~\cite{ahe2026}.
Self-Harness couples model-specific weakness discovery with proposal validation
\cite{selfharness2026}. Life-Harness~\cite{lifeharness2026}, Continual Harness~\cite{continualharness2026}, and Adaptive Auto-Harness~\cite{adaptiveautoharness2026} further extend harness evolution to runtime interventions and non-stationary task streams. MILO~\cite{jana2026miloautomatedharnessdiscovery} uses multi-agent to optimize an agent harness. However, these works focus on single agent harness evolution, leaving underexplored the scenario of refining agent harnesses in MAS with end-to-end feedback. HarnessNet closes this gap by introducing a joint harness evolution process through hierarchical feedback propagation and node-specific evolution.

\subsection{Multi-agent System (MAS) Evolution}
Recent studies study different parts of general MAS evolution. DyLAN~\cite{dylan2023} selects task-specific agent teams and adapts communication structures. GPT-Swarm~\cite{gptswarm2024} represents language agents as optimizable graphs and updates node prompts and graph connectivity, while Agentic Supernet~\cite{maas2025} searches over query-dependent multi-agent architectures. ANN~\cite{ma-etal-2026-self} combines layered execution with textual backpropagation to refine prompts and collaboration structures. MASS~\cite{zhou2026multiagentdesignoptimizingagents} coordinate the optimization of agent prompts and collaboration structures. MASPO~\cite{pmlr-v306-wang26ew} jointly refines agent prompts by evaluating their effects on downstream agents, whereas AgentGrad~\cite{chu2026agentgradinterventionguidedpromptoptimization} uses agent-level interventions to identify failure sources and guide targeted prompt updates. EvoAgentX integrates prompt and workflow optimizers~\cite{evoagentx2025}. Beyond these approaches, TPGO~\cite{he-etal-2026-learning} represents MAS textual configurations as a Textual Parameter Graph and optimizes its node content and structure. Autogenesis~\cite{zhang2026autogenesisselfevolvingagentprotocol} provides versioned resource interfaces and a closed-loop protocol for managing and evolving heterogeneous agent components. These works mostly investigated agent selection, prompt refinement and workflow search in MAS, leaving the broader harness evolution understudied. To fill this gap, HarnessNet treats the MAS harnesses as the optimization target. 

EvolveNet~\cite{nie2026evolvenetcollaborativeharnessevolution} aggregates multiple  program adaptations to evolve a shared harness across agents. In contrast, HarnessNet coordinates hierarchical local updates to execution strategies, input-output processing, and memory across heterogeneous agent harnesses.

\section{Methods}

\subsection{Agentron}

We introduce Agentron, a harness-based evolvable AI unit whose evolution is driven by its execution trajectories and downstream feedback. It is defined as
\begin{equation}
    A = (m, h, \mathcal{O}),
\end{equation}
which contains three components: \textbf{(1)} a frozen Large Language Model (LLM) $m$, which serves as the foundation model and provides reasoning capability through text; \textbf{(2)} a modifiable Forward Harness (FH) $h$, which orchestrates interactions with the model and defines the unit’s local behaviour, including context processing, tool use, and input-output handling; and \textbf{(3)} a frozen optimizer $\mathcal{O}$, an independent coding agent that receives outer feedback, analyzes execution trajectories, modifies the forward harness at the code level, and, when applicable, generates feedback for upstream units according to the code changes in forward harness and the upstream inputs as recorded in the execution trajectories.

\begin{figure}[h]
    \centering
    \includegraphics[width=\columnwidth]{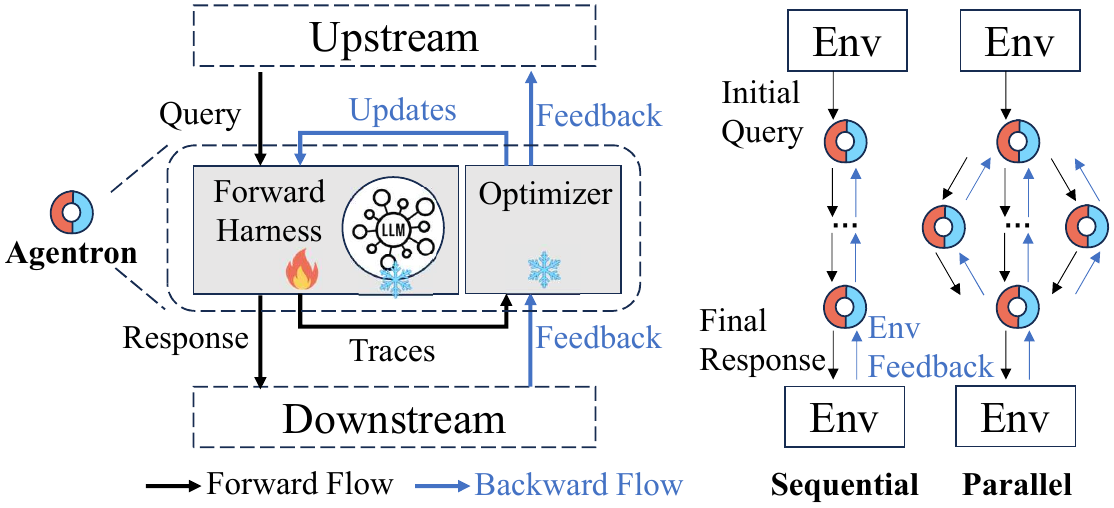}
    \caption{(Left) Agentron; (Right) Agentron Expansibility }
    \Description{Agentron and the Scalability Illustration.}
    \label{fig:agentron}
\end{figure}

As illustrated in Figure~\ref{fig:agentron}-Left, Agentron operates in two phases. In the forward phase, given a query  from upstream units, denoted by $x$, it produces a response $y$ and an execution trace $\tau$ that records evidence for subsequent analysis:
\begin{equation}
    (y,\tau) = A.\textit{Forward}(m,h,x).
\label{eq:agentron_forward}
\end{equation}
In the backward phase, given downstream feedback $s$ and the forward trace $\tau$, the frozen optimizer $\mathcal{O}$ proposes an updated harness $H'$ and generate feedbacks $f$ for upstream units when applicable, following the optimizer prompt in Appendix E.3:
\begin{equation}
\label{eq:agentron_backward}
     (h',f)=A.\textit{Backward}(\mathcal{O}, h,\tau,s).
\end{equation}

 During this two-phase process, the foundation model $M$ remains unchanged in both phases. Only forward harness $h$ is updated. An Agentron can receive messages from multiple upstream units and broadcast outputs to multiple downstream units. Its harness determines how incoming inputs are transformed and aggregated, while its optimizer can generate unit-specific feedback for upstream units. Each intermediate response feeds the downstream units as a query and local feedback as evolving signal for upstream units. This design supports building of cascade systems composed of multiple Agentron units through sequential stacking or parallel branching, with the task environment as both the starting point and the endpoint, as illustrated in Figure~\ref{fig:agentron}-Right.

\subsection{HarnessNet}

We introduce HarnessNet, a layered multi-agent system composed of multiple Agentron units. Its backbone is represented by the non-cyclic directed graph
\begin{equation}
    G = (V_1,\ldots,V_L,E),
\end{equation}
where
\begin{equation}
    V_\ell = \{v_i^\ell \mid 1 \leq i \leq N^\ell\}
\end{equation}
is the set of Agentron nodes in layer $\ell$. Each node is an Agentron of the form
\begin{equation}
    v_i^\ell =
    \left(m_i^\ell,h_i^\ell,\mathcal{O}_i^\ell\right),
\end{equation}
and
\begin{equation}
    E = \bigcup_{\ell=1}^{L-1}
        \left(V_\ell \times V_{\ell+1}\right)
\end{equation}
specifies full connectivity between adjacent layers.

\begin{figure}[!h]
    \centering
    \includegraphics[width=\columnwidth]{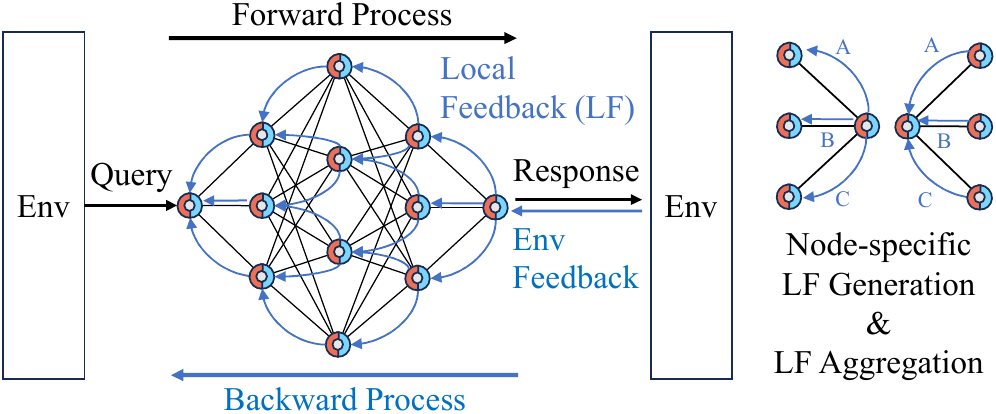}
    \caption{The dual-phase of HarnessNet (Left; some links are omitted for clarity) and the propagation of feedback (Right).}
    \Description{HarnessNet and its local feedback process.}
    \label{fig:harnessnet}
\end{figure}

As illustrated in Figure~\ref{fig:harnessnet}-Left, HarnessNet operates through a forward process and a backward process.

In the forward process \textit{(Algorithm~\ref{alg:forward})}, nodes are executed in topological order following Eq~\ref{eq:agentron_forward}. Each node receives the outputs of the connected nodes in the preceding layer. Its forward harness transforms and aggregates these messages, then it goes through a multi-round agentic reasoning process before producing and broadcasting outputs for the next layer. The first layer receives the task input from the environment, while the last layer returns the system response. Nodes in intermediate layers process information based on their roles and the behavior specified by their forward harnesses. By default, the first and last layers each contain a single node, although multiple nodes can be used to handle multiple environment inputs or outputs. Local parallelization is possible because nodes in the same layer can forward independently.

In the hierarchical backward process \textit{(Algorithm~\ref{alg:backward})}, environmental feedback is received by nodes in the last layer. Nodes are then evolved layer by layer in the reverse topological order, following Eq~\ref{eq:agentron_backward}. Each local optimizer instance updates current node forward harness after analyzing their own forward traces and local downstream feedback. Then, where applicable, feedback are generated after considering the resulting harness revisions and the upstream nodes forward outputs. This process enables node-specific local feedback generation and aggregation as illustrated in Figure~\ref{fig:harnessnet}-Right, while allowing nodes in the same layer to evolve in parallel. 

\begin{center}
\small
\begin{tabular}{@{}r p{0.92\columnwidth}@{}}
\toprule
\multicolumn{2}{@{}l@{}}{\refstepcounter{algorithm}\textbf{Algorithm \thealgorithm: HarnessNet Forward Process}\label{alg:forward}}\\
\midrule
\multicolumn{2}{@{}l@{}} {\textbf{Input:} HarnessNet backbone $G$, task query $\mathcal{Q}$}\\
\multicolumn{2}{@{}l@{}} {\textbf{Output:} Layer outputs $y$, forward trace $\mathcal{T}$}\\
1 & $y, \mathcal{T}\leftarrow\emptyset,\emptyset$\\
2 & \textbf{for} $\ell=1,\ldots,L$ \textbf{do}\\
3 & \quad $y^\ell \leftarrow \emptyset$ \\
4 & \quad $x^{\ell} \leftarrow y^{\ell-1}$ \textbf{if} $\ell>1$ \textbf{else} $\mathcal{Q}$\\
5 & \quad \textbf{for each} $v_i^\ell\in V_\ell$ \textbf{in parallel do}\\
6 & \qquad $y^\ell_i,\mathcal{T}^{\ell}_{i} \leftarrow v_i^\ell.\textit{Forward}(m_i^\ell,h_i^\ell,x^{\ell}) \qquad\qquad$ \textit{(Eq~\ref{eq:agentron_forward})}\\
7 & \qquad $y^\ell,\mathcal{T} \leftarrow y^\ell \cup y^\ell_i,\mathcal{T} \cup \mathcal{T}^{\ell}_{i}$ \\
8 & \quad \textbf{end for}\\
9 & \quad $y \leftarrow y \cup y^\ell$ \\
10 & \textbf{end for}\\
11 & \textbf{return} $(y,\mathcal{T})$ \qquad\qquad (\textit{Retrieve $y^L$ for final responses.})\\
\bottomrule
\end{tabular}
\end{center}

\begin{center}
\small
\begin{tabular}{@{}r p{0.92\columnwidth}@{}}
\toprule
\multicolumn{2}{@{}l@{}}{
\refstepcounter{algorithm}
\textbf{Algorithm \thealgorithm: HarnessNet Hierarchical Backward Process}
\label{alg:backward}
}\\
\midrule

\multicolumn{2}{@{}l@{}}{
\textbf{Input:} HarnessNet backbone $G$, forward trace $\mathcal{T}$,
}\\
\multicolumn{2}{@{}l@{}}{environment feedback $s$
}\\
\multicolumn{2}{@{}l@{}}{\textbf{Intermediate:} Upstream-layer feedback $\mathcal{F}$,}\\
\multicolumn{2}{@{}l@{}}{downstream-layer feedback $g$}\\
\multicolumn{2}{@{}l@{}}{
\textbf{Output:} Updated HarnessNet harnesses $H^{'}
$}\\

1 & $H^{'} \leftarrow \emptyset$ \\
2 & \textbf{for} $\ell=L,\ldots,1$ \textbf{do}\\
3 & \quad $\mathcal{F}^\ell \leftarrow \emptyset$\\
4 & \quad \textbf{for each} $v_i^\ell\in V_\ell$ \textbf{in parallel do}\\
5 & \qquad $(m^\ell_i,h_i^\ell,\mathcal{O}^\ell_i) \leftarrow v_i^\ell$ \\
6 & \qquad $g_i^\ell \leftarrow
\{\mathcal{F}^{\ell+1}_{k\rightarrow i} \mid v_k^{\ell+1}\in V_{\ell+1}\}$
\textbf{if} $\ell<L$ \textbf{else} $s$\\
7 & \qquad \textbf{if} $\ell>1$ \textbf{then}\\
8 & \qquad \quad
$(h_i^\ell)',\{f_{i\rightarrow j}\mid v_j^{\ell-1}\in V_{\ell-1}\}
\leftarrow v^\ell_i.\textit{Backward}($\\
  & \qquad \quad
$\mathcal{O}_i^\ell,h_i^\ell,\mathcal{T}_i^\ell,g_i^\ell) \qquad\qquad\qquad\qquad\qquad\quad$ \textit{(Eq~\ref{eq:agentron_backward})}\\
9 & \qquad \quad
$\mathcal{F}^\ell \leftarrow \mathcal{F}^\ell \cup
\{f_{i\rightarrow j}\mid v_j^{\ell-1}\in V_{\ell-1}\}$ \\
10 & \qquad \textbf{else} $(h_i^\ell)'\leftarrow
v^\ell_i.\textit{Backward}(\mathcal{O}_i^\ell,h_i^\ell,\mathcal{T}_i^\ell,g_i^\ell)\quad$ \textit{(Eq~\ref{eq:agentron_backward})} \\
11 & \qquad \textbf{end if}\\
12 & \qquad $v_i^\ell \leftarrow (m^\ell_i,(h_i^\ell)',\mathcal{O}^\ell_i)$\\
13 & \qquad $H^{'} \leftarrow H^{'} \cup (h_i^\ell)'$\\
14 & \quad \textbf{end for}\\
15 & \textbf{end for}\\
16 & \textbf{return} $H^{'}$\\
\bottomrule
\end{tabular}
\end{center}

\subsection{Task-aware HarnessNet Initialization}

Constructing a multi-agent system typically requires heavy human input about agent roles, functions, and workflow. To overcome this problem, HarnessNet uses an Initializer $\mathcal{I}$, implemented as a coding agent, to automatically plan and build initial forward harnesses for the Agentrons in the predefined HarnessNet backbone $G$. It helps automate the initialization.

The initialization process \textit{(Algorithm~\ref{alg:initialization})} takes in task-environment knowledge $K_e$, information about the nodes' reasoning backends $K_r$, and a default harness protocol $h$ specifying the repository structure, core modules, and interfaces. Using curated task-specific harness protocol $h$ is possible for more customized initialization. Actual task cases are strictly excluded from $K_e$ to prevent data leakage. 

\begin{figure}[h]
    \centering
    \includegraphics[width=\columnwidth]{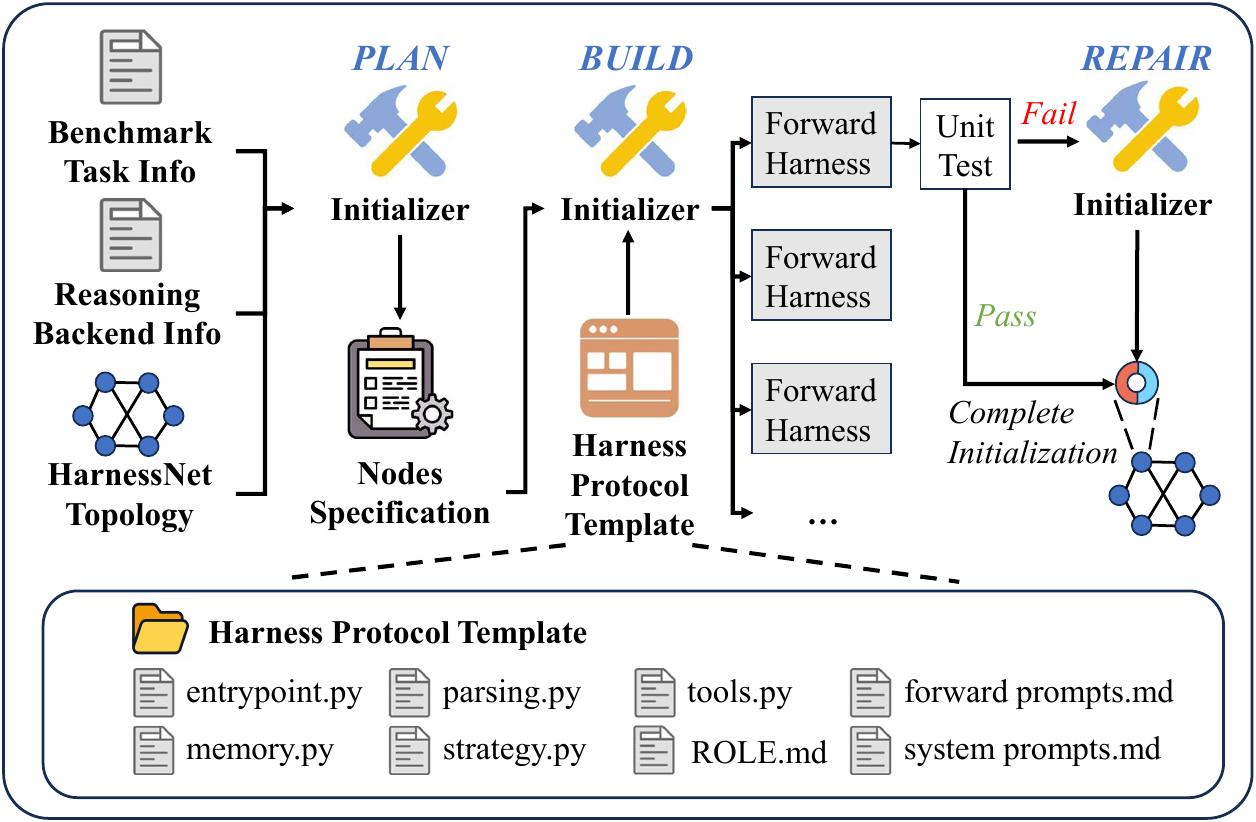}
    \caption{HarnessNet 3-Stage Task-aware Initialization Workflow. A harness template is provided at BUILD stage.}
    \Description{HarnessNet Task-aware Initialization Worlflow.}
    \label{fig:initializer}
\end{figure}

As illustrated in Figure~\ref{fig:initializer}, the Initialization proceeds in three stages. \textbf{(1)} The initializer develops a nodes specification for the network, assigning node roles and specifying strategies for memory, execution, tool use, and input-output parsing. During this process, the Initializer is instructed to assign complementary functions across nodes and avoids homogeneity. \textbf{(2)} Each node's forward harness is implemented according to this specification and the protocol $h$. \textbf{(3)} Initializer performs a smoke test on each harness and attempts a repair if the test fails. The resulting initial harnesses are assigned to the nodes of $G$ to build the HarnessNet instance. 

Prompt templates of three Initializer stages, the harness template $h$, task knowledge$K_e$ used during experiments, and reasoning backend info $K_r$  are provided in Appendix A. 

\begin{center}
\small
\begin{tabular}{@{}r p{0.92\columnwidth}@{}}
\toprule
\multicolumn{2}{@{}l@{}}{\refstepcounter{algorithm}\textbf{Algorithm \thealgorithm: HarnessNet Initialization}\label{alg:initialization}}\\
\midrule
\multicolumn{2}{@{}l@{}} {\textbf{Input:} HarnessNet backbone $G$, initializer $\mathcal{I}$, harness protocol $h$,}\\
\multicolumn{2}{@{}l@{}} {task env knowledge $K_e$, reasoning backend info $K_r$}\\
\multicolumn{2}{@{}l@{}} {\textbf{Output:} Initial HarnessNet harnesses $H^{0}$}\\
1 & $H^0\leftarrow \emptyset$\\
2 & $V_{spec}\leftarrow \mathcal{I}.Plan(G,K_e,K_r)$\\
3 & \textbf{for} $\ell=1,\ldots,L$ \textbf{do}\\
4 & \quad \textbf{for each} $v_i\in V_\ell$ \textbf{in parallel do}\\
5 & \qquad $H^l_i\leftarrow \mathcal{I}.Implement((V_{spec})^l_i,h)$\\
6 & \qquad \textbf{if} $\mathcal{I}.Test(H^l_i)$ \textbf{is} \textit{Fail} \textbf{do}\\
7 & \qquad \quad $H^l_i\leftarrow \mathcal{I}.Repair(H^l_i)$\\
8 & \qquad $H^0 \leftarrow H^0\cup H^l_i$\\
9 & \quad \textbf{end for}\\
10 & \textbf{end for}\\

11 & \textbf{return} $H^0$\\
\bottomrule
\end{tabular}
\end{center}

\subsection{HarnessNet Training}

We train HarnessNet following a batch-training-and-validate schema inspired by neural network training. This is implemented as a multi-epoch training loop. Each loop traverses the whole training set and consists of multiple training-validation steps \textit{(Algorithm~\ref{alg:validate} \& Algorithm~\ref{alg:training})}.

At each step, the training loop does the following steps: \textbf{(1)} tasks from $\mathcal{E}^{train}$ are randomly shuffled and evaluated with a fixed checkpoint $H^k$, which is the snapshot of all nodes harness, until a batch of $b$ failed cases is collected; \textbf{(2)} a backward process updates the harnesses using this batch, build the new checkpoint $H^{k+1}$; \textbf{(3)} the updated snapshot $H^{k+1}$ is evaluated on $\mathcal{E}^{val}$; \textbf{(4)} training and validation performance for this is measured by calculating the average score or case pass ratio on the training batch and validation set.

\begin{center}

\small
\begin{tabular}{@{}r p{0.92\columnwidth}@{}}
\toprule
\multicolumn{2}{@{}l@{}}{\refstepcounter{algorithm}\textbf{Algorithm \thealgorithm: HarnessNet Validation Process}\label{alg:validate}}\\
\midrule

\multicolumn{2}{@{}l@{}}{\textbf{Input:} HarnessNet backbone $G$,validation set $\mathcal{E}^{val}$}\\
\multicolumn{2}{@{}l@{}}{\textbf{Intermediates:}  Forward trace $\mathcal{T}$, layer outputs $y$, evolution step $k$,}\\
\multicolumn{2}{@{}l@{}}{ env feedback $s$, training results $\mathcal{U}$, validation results $\mathcal{V}$} \\
\multicolumn{2}{@{}l@{}}{\textbf{Output}: Validation results $\mathcal{V}$} \\

1 & $\mathcal{V} \leftarrow \emptyset$ \\
2 &  \textbf{for} $i=1,\ldots,\|\mathcal{E}^{val}\|$ \textbf{do}\\

3 &  \quad $(y_{i},\mathcal{T}_{i})
     \leftarrow G.\mathrm{\textit{Forward}}
     (\mathcal{E}^{val}_{i}.\textit{Query})$ \qquad\quad \textit{(Algorithm~\ref{alg:forward})}\\

4 &  \quad $s_{i}
     \leftarrow \mathcal{E}^{val}_i.\mathrm{\textit{Feedback}}
     (y_{i}^{L})$\\

5 &  \quad $\mathcal{V}
     \leftarrow \mathcal{V}\cup s_{i} $\\

6 & \textbf{return} $\mathcal{V}$ \\

\bottomrule
\end{tabular}
\end{center}

\begin{center}

\small
\begin{tabular}{@{}r p{0.92\columnwidth}@{}}
\toprule
\multicolumn{2}{@{}l@{}}{\refstepcounter{algorithm}\textbf{Algorithm \thealgorithm: HarnessNet Training Loop}\label{alg:training}}\\
\midrule

\multicolumn{2}{@{}l@{}}{\textbf{Input:} HarnessNet backbone $G$, failure batch size $b$,}\\
\multicolumn{2}{@{}l@{}}{training set $\mathcal{E}^{train}$, validation set $\mathcal{E}^{val}$, max training rounds $T$}\\
\multicolumn{2}{@{}l@{}}{\textbf{Intermediates:}  Forward trace $\mathcal{T}$, layer outputs $y$, }\\
\multicolumn{2}{@{}l@{}}{env feedback $s$, training results $\mathcal{U}$, validation results $\mathcal{V}$,} \\
\multicolumn{2}{@{}l@{}}{evolution step $k$, checkpoints $\mathcal{H}$} \\
\multicolumn{2}{@{}l@{}}{\textbf{Output}: Best HarnessNet harnesses checkpoint $H^\star$} \\

1 & $k,\mathcal{T},s,\mathcal{U},\mathcal{V}, \mathcal{H}
    \leftarrow 0,\emptyset,\emptyset,\emptyset,\emptyset,\emptyset$ \\
2 & $H^0 \leftarrow \{h^\ell_i|v_i^\ell \in V_{\ell},1 \leq \ell \leq L\}$  \quad \textit{(Initialized Harnesses)}\\

3 & $\mathcal{H} \leftarrow \mathcal{H} \cup H^0$ \\
4 & $\mathcal{V}^0 \leftarrow \textit{Validate(G,$\mathcal{E}^{val}$)} \qquad\qquad\qquad\quad \qquad\textit{(Algorithm~\ref{alg:validate})}$ \\

5 &  $\mathcal{V} \leftarrow \mathcal{V} \cup \mathcal{V}^0$\\

6 & \textbf{for} $t=0,\ldots,T-1$ \textbf{do}\\

7 & \quad $\mathcal{E}^{train}
    \leftarrow \mathcal{E}^{train}.\mathrm{\textit{Shuffle}}()$\\

8 & \quad $H^{k},\mathcal{U}^k \leftarrow \{h^\ell_i|v_i^\ell \in V_{\ell},1 \leq \ell \leq L\},\emptyset $ \\

9 & \quad \textbf{for} $i=1,\ldots,\|\mathcal{E}^{train}\|$ \textbf{do}\\

10 & \qquad $(y_{i},\mathcal{T}_{i})
    \leftarrow G.\mathrm{\textit{Forward}}
    (\mathcal{E}^{train}_{i}.\textit{Query})$ \quad \textit{(Algorithm~\ref{alg:forward})} \\

11 & \qquad $s_{i}
    \leftarrow \mathcal{E}^{train}_i.\mathrm{\textit{Feedback}}
    (y_{i}^{L})$\\

12 & \qquad $\mathcal{U}^k\leftarrow \mathcal{U}^k\cup s_i$ \\
13 & \qquad \textbf{if} $\mathrm{\textit{IsFailure}}(s_{i})$ \textbf{then}\\

14 & \qquad\quad $\mathcal{T}, s
     \leftarrow
     \mathcal{T} \cup \mathcal{T}_{i}, s\cup s_i$\\

15 & \qquad\quad \textbf{if} $\|s\|=b$ \textbf{or} $i==\|\mathcal{E}^{train}\|$ \textbf{then} \\

16 & \qquad\qquad $H^{k+1}
     \leftarrow G.\mathrm{\textit{Backward}}
     (\mathcal{T}, s)$\qquad\quad \textit{(Algorithm~\ref{alg:backward})}\\

17 & \qquad \quad \quad $\mathcal{V}^{k+1} \leftarrow \textit{Validate(G,$\mathcal{E}^{val}$)} \qquad \qquad\textit{(Algorithm~\ref{alg:validate})}$\\
18 & \qquad \quad \quad $\mathcal{U},\mathcal{V}, \mathcal{H} \leftarrow \mathcal{U} \cup \mathcal{U}^k,\mathcal{V} \cup \mathcal{V}^{k+1}, \mathcal{H} \cup H^{k+1}$ \\

19 & \qquad\qquad $\mathcal{T},s,\mathcal{U}^k,\mathcal{V}^{k+1}
    \leftarrow \emptyset,\emptyset,\emptyset,\emptyset$\\
20 & \qquad\qquad $k \leftarrow k+1$ \\
21 & \qquad\quad \textbf{end if}\\

22 & \qquad \textbf{end if}\\

23 & \quad \textbf{end for}\\

24 & \textbf{end for}\\

25 & $H^\star
     \leftarrow \mathrm{\textit{GetBest}}
     (\mathcal{H},(\mathcal{U,V}))$\\

26 & \textbf{return} $H^\star$ \qquad\qquad\qquad \textit{(Trained HarnessNet Harnesses)}\\

\bottomrule
\end{tabular}
\end{center}

The final snapshot $H^\star$ is selected from all evaluated snapshots. Similar to traditional neural network training, we select the best training checkpoint $H^\star$ by considering both the training performance $\mathcal{U}$ and validation performance $\mathcal{V}$, based on the performance curve convergence status and point scores. 

In order to reduce the experiment cost, at each training step, we only use failed cases to reduce the number of backward processes needed to traverse the $\mathcal{E}^{train}$. But it is allowed to include successful cases and might help improve the stability and quality of evolution process according to some recent works~\citep{chen2026skillcat,jiang2026harnessevolve,kang2026acon}. 

\section{Experiments}

\subsection{Experiment Settings}
\subsubsection{Dataset}

We use AgentBench~\cite{liu2024agentbench}, a light-weight agentic benchmark, to demonstrate the training feasibility of proposed Agentron and HarnessNet. We focus on ALFWorld and WebShop, whose training splits reuse the original benchmarks’ official training data, while other domains rely on synthetic training data, which may introduce train–test distribution shifts~\cite{zhang2025agentrlscalingagenticreinforcement}. To save training efforts while preventing testset leakage, we created  training and validation set using the randomly sampled cases from \textit{alfworld-env\_train} and \textit{webshop-env\_train}, and then test the selected harness checkpoints on \textit{alfworld-std} and \textit{webshop-std}, the official benchmark testsets. The number of cases in different splits are shown in Table~\ref{tab:datasplits}. The environment task feedback is a binary case pass/fail label in ALFWorld, and a numeric 0-1 case score in WebShop.

\begin{table}[!h]
\caption{Experiment Data Splits}
\centering
\small
\setlength{\tabcolsep}{2pt}
\begin{tabular}{@{}lcccccc@{}}
\toprule
\textbf{Benchmark}                     & \multicolumn{3}{c}{ALFWorld}                   & \multicolumn{3}{c}{WebShop} \\ \midrule
\multicolumn{1}{l|}{\textbf{Split}}    & Train & Validation & \multicolumn{1}{c|}{Test} & Train  & Validation  & Test \\
\multicolumn{1}{l|}{\textbf{Case Num}} & 168   & 42         & \multicolumn{1}{c|}{109}  & 320    & 80          & 200  \\ \bottomrule
\end{tabular}
\label{tab:datasplits}
\end{table}

\subsubsection{Foundation Models, Initializer and Optimizer}
To simplify the setup and reduce computational costs, in the experiments, we equip Agentrons in the same HarnessNet with independent instances of the same foundation model and optimizer. Models include Qwen3.5-9B~\cite{qwen3.5}, Gemma4-8b-e4b~\cite{team2026gemma}, Ornith 1.5-9b~\cite{ornith_1_5} and Qwen3.8-27B~\cite{qwen38}, using ollama~\cite{ollama} as reasoning backend. We use GPT5.6-SOL High~\cite{openai_gpt56} combined with codex-cli 0.153.4 as both the Initializer and Optimizer. We use zero temperature and fixed dataset sampling/shuffling seeds to improve reproducibility.  Complete model hyperparameter settings and information are provided in Appendix B.

\subsection{Training Effectiveness}
We assess training feasibility for Agentron and HarnessNet in five steps: \textbf{(1)} Evaluate foundation models on the test sets to establish the Model-only baseline. \textbf{(2)} Instantiate three minimal topologies and initialize their harnesses using Algorithm~\ref{alg:initialization}. \textbf{(3)} Measure pre-training test performance (FH0). \textbf{(4)} Train each system using Algorithm~\ref{alg:training}, assessing convergence and guiding stopping with training and validation curves. \textbf{(5)} Test the selected forward harness checkpoints (FH\_starred). We compute \textit{Harness Gain} relative to Model-only and \textit{Training Gain} relative to FH0. Results show that feedback-driven iterative training can improve system performance on unseen task cases. Detailed training records are in Appendix C.

\begin{figure*}[!t]
    \centering
    \includegraphics[width=\textwidth]{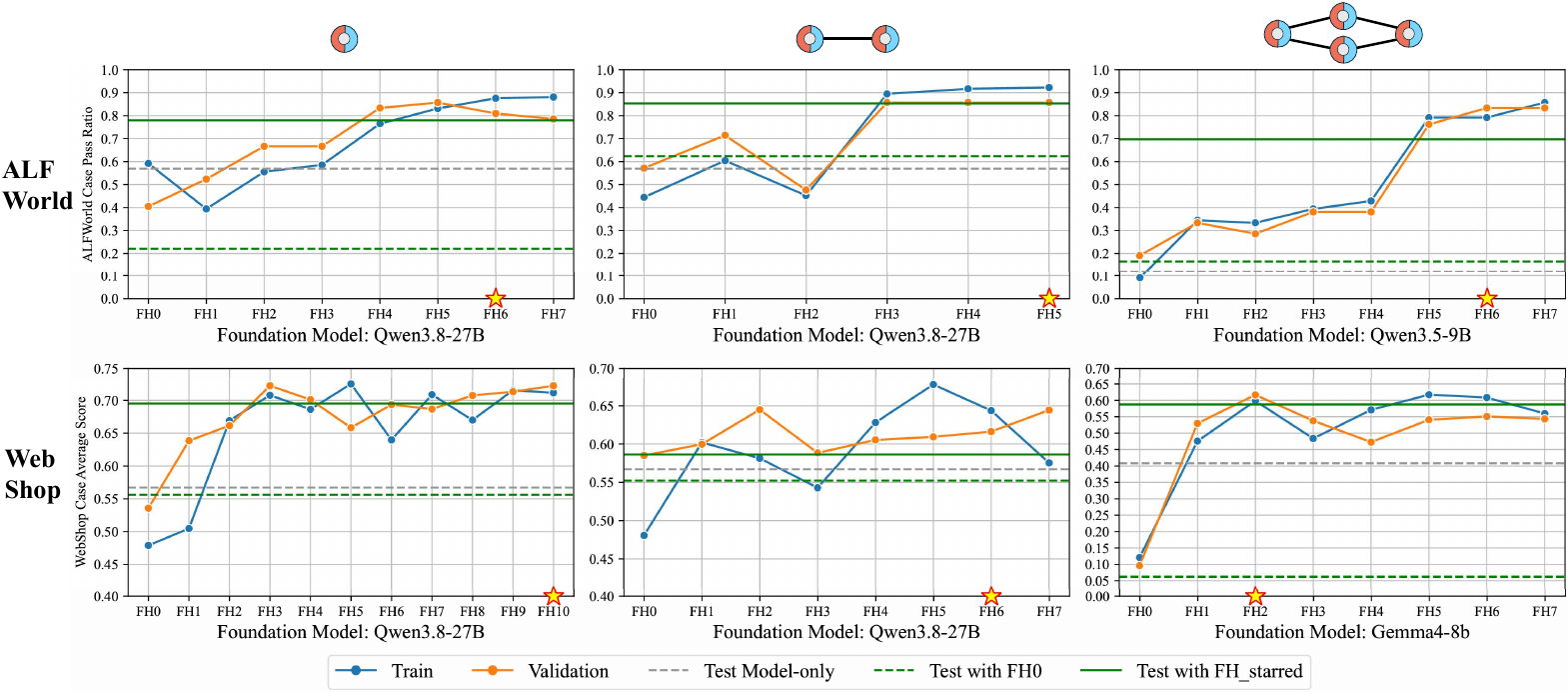}
    \caption{Training Curves and Test Results on AgentBench of Minimal HarnessNet Topologies. }
    \label{fig:training}
\end{figure*}

\subsubsection{Training Agentron}
We firstly verify the training feasibility of the fundamental unit, Agentron, which is also a single-node HarnessNet. As the first column in Figure~\ref{fig:training} shows, the training process quickly improves Agentron performance and converges even though the initialized FH0s were bad (FH0 test score $<$ Model-only baseline). Results are in Table~\ref{tab:training_agentron} and detailed training records are in Appendix C.1-2. 

\begin{table}[H]
\caption{Agentron Training Effects}

\centering
\small
\setlength{\tabcolsep}{2pt}
\begin{tabular}{@{}lllll@{}}
\toprule
\textbf{\begin{tabular}[c]{@{}l@{}}Benchmark\\ Domain\end{tabular}} & \textbf{Test} & \textbf{\begin{tabular}[c]{@{}l@{}}Score\\ (↑)\end{tabular}} & \textbf{\begin{tabular}[c]{@{}l@{}}Harness\\ Gain\end{tabular}} & \textbf{\begin{tabular}[c]{@{}l@{}}Training\\ Gain\end{tabular}} \\ \midrule
ALF-World                                          & Model-only    & 0.5688                                                       & -                                                               & -                                                                \\
                                                                    & FH0           & 0.2202                                                       & -61.29\%                                                        & -                                                                \\
                                                                    & FH\_starred   & 0.7798                                                       & +37.10\%                                                        & +254.13\%                                                        \\ \midrule
WebShop                                            & Model-only    & 0.5670                                                       & -                                                               & -                                                                \\
                                                                    & FH0           & 0.5559                                                       & -1.96\%                                                         & -                                                                \\
                                                                    & FH\_starred   & 0.6950                                                       & +22.57\%                                                        & +25.02\%                                                         \\ \bottomrule
\end{tabular}

\label{tab:training_agentron}
\end{table}

\subsubsection{Training with Minimal Cross-layer Feedback}

After verifying Agentron's training, we build the minimal 2-node harnessnet to check the training feasibility when the cross-layer feedback propagation is enabled. As the second column in Figure~\ref{fig:training} shows, the FH0 is well initialized with close performance to Model-only baseline. Subsequent training further improves the performance. Test results are in Table~\ref{tab:training_2node} and detailed training records are in Appendix C.3-4.

\begin{table}[h]
\caption{2-Node HarnessNet Training Effects}
\centering
\small
\setlength{\tabcolsep}{2pt}
\begin{tabular}{@{}lllll@{}}
\toprule
\textbf{\begin{tabular}[c]{@{}l@{}}Benchmark\\ Domain\end{tabular}} & \textbf{Test} & \textbf{\begin{tabular}[c]{@{}l@{}}Score\\ (↑)\end{tabular}} & \textbf{\begin{tabular}[c]{@{}l@{}}Harness\\ Gain\end{tabular}} & \textbf{\begin{tabular}[c]{@{}l@{}}Training\\ Gain\end{tabular}} \\ \midrule
ALF-World                                          & Model-only    & 0.5688                                                       & -                                                               & -                                                                \\
                                                                    & FH0           & 0.6239                                                       & +9.69\%                                                         & -                                                                \\
                                                                    & FH\_starred   & 0.8532                                                       & +50.00\%                                                        & +36.75\%                                                         \\ \midrule
WebShop                                            & Model-only    & 0.5670                                                       & -                                                               & -                                                                \\
                                                                    & FH0           & 0.5523                                                       & -2.59\%                                                         & -                                                                \\
                                                                    & FH\_starred   & 0.5892                                                       & +3.92\%                                                         & +6.68\%                                                          \\ \bottomrule
\end{tabular}
\label{tab:training_2node}
\end{table}

\subsubsection{Training with Minimal Node-specific Feedback Generation \& Aggregation}
Finally, we evaluate a 4-node HarnessNet with an input node, an output node, and two parallel hidden nodes to assess training feasibility with cross-layer, node-specific feedback generation and aggregation mechanism. To reduce computational costs, we use the smaller Qwen3.5-9B~\cite{qwen3.5} and Gemma4-8b-e4b~\cite{team2026gemma} models. As shown in the third column of Figure~\ref{fig:training}, Qwen3.5-9B fails 88.07\% ALFWorld test cases in the Model-only setting. However, the 4-node HarnessNet starts from a beneficial FH0 and, through training, achieves performance comparable to that obtained with Qwen3.8-27B. On WebShop, training improves the Gemma4-8b-e4b system from a poorly performing FH0 to a checkpoint that surpasses the Qwen3.8-27B Model-only baseline and matches 2-node HarnessNet training performance. Test results are in Table~\ref{tab:training_4node} and details are in Appendix C.5-6.

\begin{table}[h]
\caption{4-Node HarnessNet Training Effects}

\centering
\small
\setlength{\tabcolsep}{2pt}
\begin{tabular}{@{}lllll@{}}
\toprule
\textbf{\begin{tabular}[c]{@{}l@{}}Benchmark\\ Domain\end{tabular}} & \textbf{Test} & \textbf{\begin{tabular}[c]{@{}l@{}}Score\\ (↑)\end{tabular}} & \textbf{\begin{tabular}[c]{@{}l@{}}Harness\\ Gain\end{tabular}} & \textbf{\begin{tabular}[c]{@{}l@{}}Training\\ Gain\end{tabular}} \\ \midrule
ALF-World                                          & Model-only    & 0.1193                                                       & -                                                               & -                                                                \\
                                                                    & FH0           & 0.1651                                                       & +38.39\%                                                        & -                                                                \\
                                                                    & FH\_starred   & 0.6972                                                       & +484.41\%                                                       & +322.29\%                                                        \\ \midrule
WebShop                                            & Model-only    & 0.4083                                                       & -                                                               & -                                                                \\
                                                                    & FH0           & 0.0624                                                       & -84.72\%                                                        & -                                                                \\
                                                                    & FH\_starred   & 0.5875                                                       & +43.89\%                                                        & +841.51\%                                                        \\ \bottomrule
\end{tabular}

\label{tab:training_4node}
\end{table}

\begin{figure*}[!t]
    \centering

    \includegraphics[width=\textwidth]
        {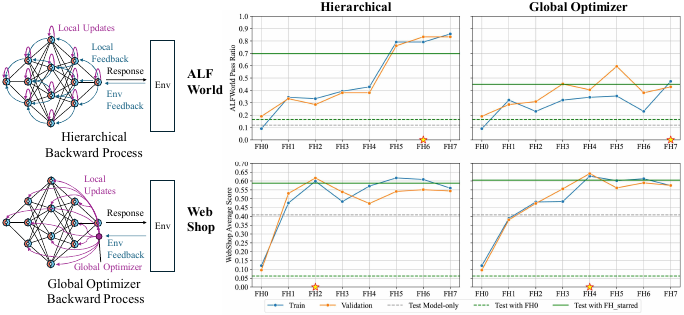}
    \caption{(Left) Illustration of Hierarchical Approach
        \& Global Optimizer Approach; (Right) Training Performance.}
    \label{fig:ablation_performance}

    \vspace{0.5em}

    \includegraphics[width=\textwidth]
        {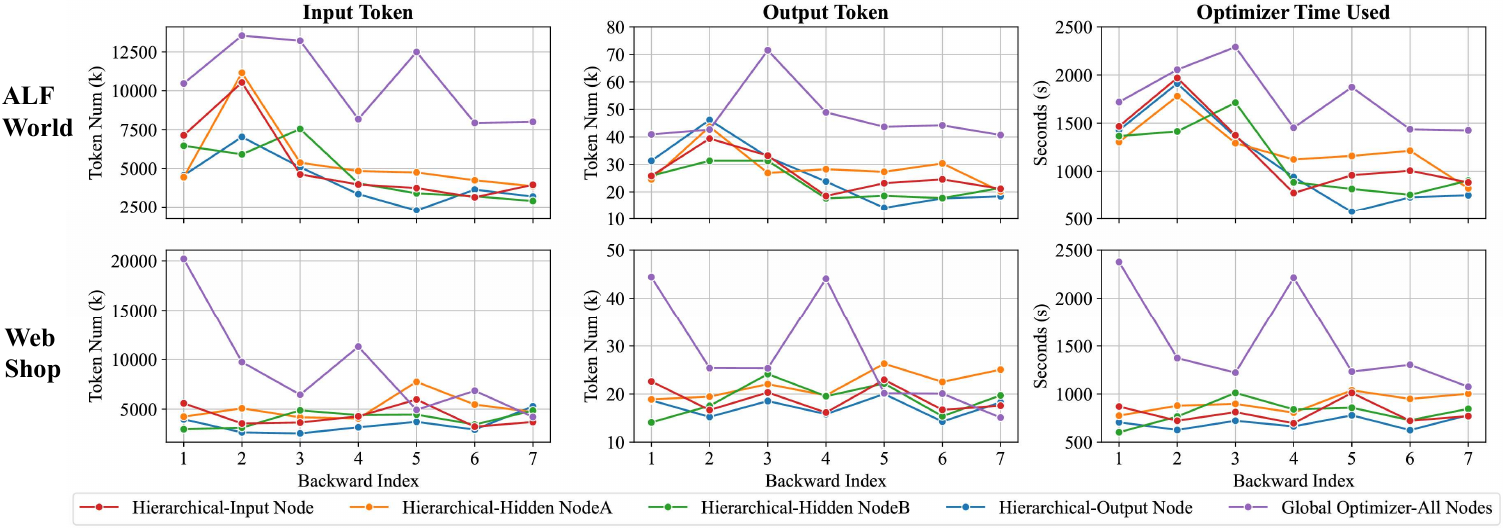}
    \caption{Per-call Optimizer Cost and Latency of Hierarchical Approach
        vs. Global Optimizer Approach.}
    \label{fig:ablation_cost}
\end{figure*}

\subsection{Hierarchical Approach vs. Global Optimizer}

To evaluate whether hierarchical evolution reduces the burden on single optimizer-call while maintaining competitive evolution effect, we compare it with the global optimizer approach \textit{(Algorithm~\ref{alg:backward_flat})} on the 4-node HarnessNet topology. Global optimizer prompt is provided in Appendix E.3.

\begin{center}
\small
\begin{tabular}{@{}r p{0.92\columnwidth}@{}}
\toprule
\multicolumn{2}{@{}l@{}}{
\refstepcounter{algorithm}
\textbf{Algorithm \thealgorithm: Global Optimizer Backward Process}
\label{alg:backward_flat}
}\\
\midrule

\multicolumn{2}{@{}l@{}}{
\textbf{Input:} HarnessNet backbone $G$, forward trace $\mathcal{T}$, 
}\\
\multicolumn{2}{@{}l@{}}{
environment feedback $s$
}\\
\multicolumn{2}{@{}l@{}}{
\textbf{Intermediate:} Global harness set $\mathcal{H}$, update set $\mathcal{U}$
}\\
\multicolumn{2}{@{}l@{}}{
\textbf{Output:} Updated harnesses $H^{'}$
}\\

1 & $H \leftarrow
\{h^\ell_i|v_i^\ell \in V_{\ell},1 \leq \ell \leq L\}$ \\

2 & $H^{'} \leftarrow
\mathcal{O}_{\mathrm{global}}.\textit{Backward}
(G,H,\mathcal{T},s)$ \\

3 & \textbf{for each}
$(h_i^\ell)'\in H^{'} $ \textbf{do}\\

4 & \quad $(m_i^\ell,h_i^\ell,\mathcal{O}_i^\ell)
\leftarrow v_i^\ell$\\

5 & \quad
$v_i^\ell \leftarrow
(m_i^\ell,(h_i^\ell)',\mathcal{O}_i^\ell)$\\

6 & \textbf{end for}\\

7 & \textbf{return} $H^{'}$\\

\bottomrule
\end{tabular}
\end{center}

 Both methods start from the same initialized FH0 and the first batch of failed trajectories, and undergo the same number of backward steps on identical training and validation sets. Finally, We compare the checkpoint performances and per-optimizer-call costs. Detailed results are in Appendix E.

 Figure~\ref{fig:ablation_performance}-Left illustrates two approaches. In the hierarchical backward process, each Agentron uses its local optimizer to update its forward harness based on node-specific feedback from the downstream layer. By contrast, the global optimizer approach uses a centralized global optimizer, implemented as an independent optimizer instance while disabling other Agentron local optimizers. It is granted with access to all nodes forward harnesses and forward trajectories to jointly update the entire system based on environment feedback.

As Figure~\ref{fig:ablation_performance}-Right shows, both approaches improve over initialization. Within the same number of backward steps, hierarchical evolution process achieves comparable performance on WebShop and substantially higher performance on ALFWorld, suggesting its practicability and evolving effectiveness. Detailed records are provided in Appendix E.1.

Figure~\ref{fig:ablation_cost} compares the per-optimizer-call cost of these two approaches. Although the total end-to-end cost may increase due to the fixed cost in each optimizer call, such as repeatedly loading prompt prefixes and CPU process-switching latency, individual local optimizer calls require fewer input and output tokens and less execution time than global optimizer calls. These results indicate that hierarchical optimization reduces per-optimizer context and workload burdens while maintaining competitive system-level improvement. Detailed cost data records are provided in Appendix E.2.

\subsection{Ability Migration}

We assess harness transferability by replacing Qwen3.5-9B~\cite{qwen3.5} in the trained 4-node ALFWorld HarnessNet with either Ornith1.5-9B~\cite{ornith_1_5}, an open-source model fine-tuned from Qwen3.5-9B, or Gemma4-8b-e4b~\cite{team2026gemma}, an open-source model from a different family. We evaluate the migrated harness checkpoints on the same validation set during training. 
\begin{figure}[h]
    \centering
    \begin{subfigure}{\columnwidth}
        \centering
        \includegraphics[width=\columnwidth]{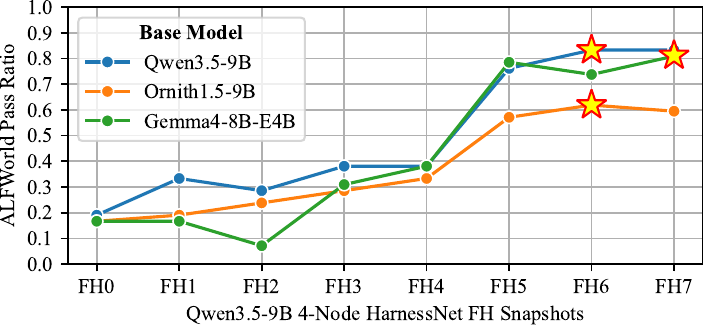}
        \Description{Agentron training results on ALFWorld with Qwen3.5-9B as the frozen foundation model.}
        \label{fig:migrate-validation}
    \end{subfigure}
    \begin{subfigure}{\columnwidth}
        \centering
        \includegraphics[width=\columnwidth]{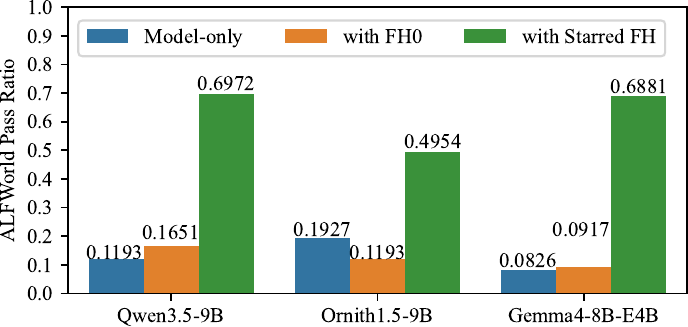}
        \Description{4-node HarnessNet migration results on ALFWorld validation with different frozen foundation models.}
        \label{fig:harnessnet-migration}
    \end{subfigure}

    \caption{Validation (Up) and Test Scores (Bottom) on ALFWorld across Different Foundation Models and Identical 4-node HarnessNet FH Snapshots Trained with Qwen3.5-9B.}
    \Description{Agentron training results with Qwen3.5-9B and 4-node HarnessNet migration scores on the ALFWorld validation set.}
    \label{fig:alfworld-training-migration}
\end{figure}

As shown in the upper panel of Figure~\ref{fig:alfworld-training-migration}, both replacement models benefit from the harnesses evolved with Qwen3.5-9B and exhibit similar performance improvement trends across checkpoints. Their validation performance curves exhibit similar trend with the original model. We then select FH\_starred based on validation performance and evaluate it on the same test set. All three models fail most test case in the Model-only setting. When equipped with the same FH0, they both reach the similar test performance as Qwen3.5-9B. With the evolved 4-node harnesses, however, Gemma4-8b-e4b and Ornith1.5-9B achieve success rates of 68.81\% and 49.54\%, which coheres with the results observed with Qwen3.5-9B. These results indicate that task-specific mechanisms learned within individual HarnessNet harnesses remain jointly effective across foundation models, including models from a different family. Detailed validation and test records are provided in Appendix D, with involved model details and hyperparameter settings provided in Appendix B.

\subsection{Case Study - Learn Rules, Not Answers}

In Appendix F, we present a case study using training experiment records of 4-node HarnessNet. It shows that during the hierarchical harness evolution process, MAS can learn useful rules from end-to-end task-environment feedback rather than merely memorizing answers. The study also shows that Harness-based evolution, an emerging trend in the evolution of AI systems, occurs in a human-friendly modality space (text space) and is characterized by good auditability. This supports efforts to improve the transparency and controllability of modern AI system evolution, which builds convenient interface for potential human intervention in the evolved agent behaviors and mechanisms.

\section{Future Work}
Due to the limit of computation resources, we propose HarnessNet but only validated its training feasibility on some minimal cases. The future directions point to validate this feedback-driven hierarchical evolution framework on larger scale, which includes: \textbf{(1)} validating training feasibility on more complicated topology consists of more and wider layers; \textbf{(2)} equipping each Agentron with larger parameter volume LLMs; \textbf{(3)} training on more heavy-weight and complex agentic benchmark tasks; \textbf{(4)} testing with a broader variety of optimizers, especially some open-sourced coding agents.

\section{Conclusion}
We propose HarnessNet and its fundamental unit, Agentron, for feedback-driven hierarchical joint harness evolution in multi-agent systems. Experiments and studies show: \textbf{(1)} iterative training on some minimal cases is feasible and can improves the task performance based on end-to-end task feedback; \textbf{(2)} evolved harnesses capabilities are transferrable across foundation models of similar size; and \textbf{(3)} compared with single global optimizer approach, hierarchical evolution approach reduces per-optimizer-call workload and context burdens while keeping comparable training effects under constraint local optimizer access. Our work provides a promising solution for harness-based MAS evolution. The auditability coheres with the trend to elicit AI systems evolution in a human-friendly modality space.

\clearpage

\bibliography{references}

\onecolumn
\appendix
\setcounter{secnumdepth}{2}

\renewcommand{\thesection}{Appendix~\Alph{section}}
\renewcommand{\thesubsection}{\Alph{section}.\arabic{subsection}}
\renewcommand{\thetable}{\Alph{section}\arabic{table}}
\renewcommand{\thefigure}{\Alph{section}\arabic{figure}}
\renewcommand{\thelstlisting}{\Alph{section}\arabic{lstlisting}}
\makeatletter
\@addtoreset{table}{section}
\@addtoreset{lstlisting}{section}
\makeatother
\setcounter{table}{0}
\setcounter{figure}{0}
\setcounter{lstlisting}{0}

\begin{center}
  \vspace*{0.12\textheight}
  {\LARGE\bfseries HarnessNet: Supplementary Material}

  \vspace{2.5em}
  {\large\bfseries Appendix Navigation}
\end{center}

\vspace{1em}
\begingroup
\large
\renewcommand{\arraystretch}{1.55}
\noindent\begin{tabularx}{\textwidth}{@{}>{\raggedright\arraybackslash}X@{}}
  \toprule
  \hyperref[app:initializer]{\textbf{\ref*{app:initializer}}\enspace Initializer Implementation Details} \\
  \hyperref[app:model-configurations]{\textbf{\ref*{app:model-configurations}}\enspace Forward-Model Configurations} \\
  \hyperref[app:training-experiments]{\textbf{\ref*{app:training-experiments}}\enspace Training Experiments} \\
  \hyperref[app:migration-study]{\textbf{\ref*{app:migration-study}}\enspace Migration Study} \\
  \hyperref[app:optimizer-ablation]{\textbf{\ref*{app:optimizer-ablation}}\enspace Hierarchical vs. Global Optimizer Comparison Study Details} \\
  \hyperref[app:case-study]{\textbf{\ref*{app:case-study}}\enspace Case Study: Learn Rules, Not Answers} \\
  \bottomrule
\end{tabularx}
\endgroup
\vfill
\clearpage

\section{Initializer Implementation Details}
\label{app:initializer}

The HarnessNet initializer constructs the trainable forward-harness baseline
(FH0) before any benchmark case is used for optimization.  It is implemented as
an auditable controller protocol with three model-facing stages: \textsc{Plan},
\textsc{Build}, and \textsc{Repair}.  The controller, rather than the model,
owns job publication, state transitions, validation, and final acceptance.  The
prompt templates below retain the implementation wording, with only line wrapping
and continuation indentation adjusted for typesetting.

\noindent\textbf{Subsection navigation}
\par\smallskip
\begingroup
\small
\setlength{\tabcolsep}{8pt}
\renewcommand{\arraystretch}{1.08}
\noindent\begin{tabularx}{\textwidth}{@{}>{\raggedright\arraybackslash}X@{}}
  \toprule
  \hyperref[app:initializer-inputs]{\ref*{app:initializer-inputs}\enspace Frozen inputs and visibility boundary} \\
  \hyperref[app:initializer-environment-backend]{\ref*{app:initializer-environment-backend}\enspace Task-environment and reasoning-backend information} \\
  \hyperref[app:initializer-protocol]{\ref*{app:initializer-protocol}\enspace Three-stage protocol} \\
  \hyperref[app:initializer-acceptance]{\ref*{app:initializer-acceptance}\enspace Execution and acceptance separation} \\
  \hyperref[app:initializer-plan-prompt]{\ref*{app:initializer-plan-prompt}\enspace PLAN prompt template} \\
  \hyperref[app:initializer-build-prompt]{\ref*{app:initializer-build-prompt}\enspace BUILD prompt template} \\
  \hyperref[app:initializer-repair-prompt]{\ref*{app:initializer-repair-prompt}\enspace REPAIR prompt template} \\
  \hyperref[app:initializer-harness-template]{\ref*{app:initializer-harness-template}\enspace Canonical harness template} \\
  \bottomrule
\end{tabularx}
\endgroup
\medskip

\subsection{Frozen inputs and visibility boundary}
\label{app:initializer-inputs}

Before initialization starts, the controller freezes the public benchmark
profile, LLM-backend profile, topology, harness contract, canonical template
manifest, SDK guide, and model controls.  The public benchmark profile may
describe the task family, action or output schema, scoring semantics, and
available tools, but it does not expose selected cases, case identifiers,
observations, answers, or case-level statistics.  This boundary prevents the
initializer from encoding benchmark instances into FH0.

Each agent is materialized as an independent Git repository from the same
canonical harness template.  The topology fixes node identities, layer
positions, edges, and input and output boundary roles.  The stable Contract 1.2
runtime supplies one backend call per node invocation, preserves the complete
case-local native-message history, structurally projects a single native tool
call when present, and propagates execution errors through a stable error
envelope.

\subsection{Task-environment and reasoning-backend information}
\label{app:initializer-environment-backend}

FH0 initialization consumes reusable public control inputs rather than benchmark
cases.  A benchmark-specific \texttt{benchmark\_profile.json} describes the task
environment, interaction protocol, tools, output contract, termination, and
scoring rules.  The shared \texttt{llm\_backend\_profile.json} is a concrete
runtime profile rather than a generic backend description: it specifies the
backend type, model identifier, generation controls, call limits, and retry
policy.  In the initializer workflow, PLAN and BUILD can read the full backend
profile.  REPAIR receives the accepted initialization plan, validation evidence,
and benchmark profile, but the backend profile is not injected again as an
independent input.

Listings~\ref{lst:initializer-alfworld-task-environment} and
\ref{lst:initializer-webshop-task-environment} provide the complete public
task-environment profiles for ALFWorld and WebShop, respectively.
Listing~\ref{lst:initializer-reasoning-backend} provides the shared reasoning-
backend profile; its local Ollama registry ID is replaced by the normalized paper
name \texttt{Qwen3.8-27b}.

\begin{lstlisting}[
  style=prompttemplate,
  caption={ALFWorld task-environment information supplied to FH0 initialization.},
  label={lst:initializer-alfworld-task-environment}
]
{
  "adapter": "agentbench-alfworld",
  "benchmark_family": "ALFWorld",
  "category_semantics": {
    "look_at_obj": "pick up the target object and examine it using the requested light source",
    "pick_and_place": "pick up one target object and place it in or on the requested receptacle",
    "pick_clean_then_place": "clean one target object and then place it in or on the requested receptacle",
    "pick_cool_then_place": "cool one target object and then place it in or on the requested receptacle",
    "pick_heat_then_place": "heat one target object and then place it in or on the requested receptacle",
    "pick_two_obj": "pick up two instances of the target object and place both in or on the requested receptacle"
  },
  "initializer_visibility": {
    "allowed": "public task, interaction, tool, scoring, and category semantics only",
    "case_data_access": false
  },
  "interaction": {
    "action_policy": "choose exactly one action from the current admissible_actions",
    "action_tool": {
      "arguments_schema": {
        "additionalProperties": false,
        "properties": {
          "action": {
            "minLength": 1,
            "type": "string"
          }
        },
        "required": [
          "action"
        ],
        "type": "object"
      },
      "name": "take_action",
      "owner": "agentbench_adapter",
      "required": true
    },
    "initializer_contract_probe": {
      "admissible_action": "look",
      "observation": "Here is your task. Put a synthetic object in a synthetic receptacle. You are in a synthetic room. AVAILABLE ACTIONS: look\ninventory\ngo to synthetic receptacle 1\n",
      "synthetic": true
    },
    "invalid_turns": {
      "invalid_action": {
        "feedback": "the turn is rejected and the agent is asked for a valid tool action",
        "may_continue": true,
        "reward": 0
      },
      "missing_tool_call": {
        "feedback": "the turn is rejected and the agent is asked to call the action tool",
        "may_continue": true,
        "reward": 0
      }
    },
    "mode": "multi_step",
    "observation_contains": [
      "task_goal",
      "environment_feedback",
      "admissible_actions"
    ],
    "observation_transport": {
      "content_type": "text/plain",
      "encoding": "one raw text string in AgentBench ALFWorld format; task goal, feedback, and AVAILABLE ACTIONS are not pre-parsed request fields",
      "request_path": "upstream_inputs[0].content",
      "source_agent_id": "environment"
    },
    "repeated_action_termination": {
      "consecutive_limit": 3,
      "terminal_status": "agent invalid action"
    },
    "step_feedback": {
      "admissible_actions_refreshed": true,
      "environment_observation_available": true,
      "reward_meaning": "the environment reports whether the household goal has been completed"
    },
    "termination": {
      "environment_done": "the environment ends the interaction when its task reaches a terminal state",
      "step_limit": 15,
      "step_limit_status": "task limit reached"
    }
  },
  "network_output_contract": {
    "accepted_last_content": [
      "action string",
      {
        "action": "action string"
      }
    ],
    "adapter_encoding": "the AgentBench adapter wraps the last-node action in take_action",
    "intermediate_nodes": "emit useful structured content for downstream HarnessNet nodes",
    "last_node": "emit exactly one chosen admissible environment action"
  },
  "schema_version": "1.0",
  "scoring": {
    "aggregate_definition": "successful completed tasks divided by evaluated tasks",
    "case_result_field": "result.result",
    "case_success": "result.result == 1",
    "failure_value": 0,
    "primary_metric": "success_rate",
    "success_value": 1
  },
  "task_categories": [
    "pick_and_place",
    "pick_clean_then_place",
    "pick_heat_then_place",
    "pick_cool_then_place",
    "look_at_obj",
    "pick_two_obj"
  ],
  "task_description": {
    "agent_role": "read the current goal and environment feedback, then choose the next valid environment action",
    "objective": "complete the stated household manipulation goal through a sequence of environment actions",
    "setting": "interactive household environment"
  },
  "test_split": "new_std",
  "test_task": "alfworld-std",
  "train_task": "alfworld-env_train"
}
\end{lstlisting}

\begin{lstlisting}[
  style=prompttemplate,
  caption={WebShop task-environment information supplied to FH0 initialization.},
  label={lst:initializer-webshop-task-environment}
]
{
  "adapter": "agentbench-webshop",
  "benchmark_family": "WebShop",
  "initializer_visibility": {
    "allowed": "public task family, interaction, tool, scoring, topology, harness contract, template, and backend controls only",
    "case_data_access": false
  },
  "interaction": {
    "action_policy": "emit exactly one published WebShop tool with an exact argument object valid for the current page",
    "initializer_contract_probe": {
      "expected_output": {
        "arguments": {
          "keywords": "synthetic red mug"
        },
        "tool_name": "search_action"
      },
      "observation": "Synthetic WebShop task: search for a synthetic red mug. Available Actions: search.",
      "synthetic": true
    },
    "invalid_turns": {
      "invalid_arguments": {
        "feedback": "the controller ignores arguments that do not match the selected tool",
        "may_continue": true,
        "reward": 0
      },
      "invalid_tool": {
        "feedback": "the controller ignores an unknown tool name",
        "may_continue": true,
        "reward": 0
      },
      "missing_tool_call": {
        "feedback": "the controller requests one executable WebShop tool call",
        "may_continue": true,
        "reward": 0
      }
    },
    "mode": "multi_step",
    "observation_transport": {
      "content_type": "text/plain",
      "encoding": "the initial observation contains the shopping instruction and available actions; later observations contain the current page and updated available actions",
      "request_path": "upstream_inputs[0].content",
      "source_agent_id": "environment"
    },
    "step_feedback": {
      "environment_observation_available": true,
      "reward_meaning": "the controller reports a continuous shopping-goal match reward from zero through one"
    },
    "termination": {
      "environment_done": "the controller ends the interaction after purchase or another terminal page state",
      "step_limit": 20,
      "step_limit_status": "task limit reached"
    },
    "tools": [
      {
        "arguments_schema": {
          "additionalProperties": false,
          "properties": {
            "keywords": {
              "type": "string"
            }
          },
          "required": [
            "keywords"
          ],
          "type": "object"
        },
        "name": "search_action",
        "owner": "agentbench_adapter",
        "required": false
      },
      {
        "arguments_schema": {
          "additionalProperties": false,
          "properties": {
            "value": {
              "type": "string"
            }
          },
          "required": [
            "value"
          ],
          "type": "object"
        },
        "name": "click_action",
        "owner": "agentbench_adapter",
        "required": false
      }
    ]
  },
  "network_output_contract": {
    "adapter_encoding": "the AgentBench adapter validates the object and submits the matching native tool-call envelope",
    "last_node": "emit one object containing tool_name and arguments",
    "memory": "case-local harness memory only",
    "single_node_role": "the sole node is both the first and last HarnessNet node"
  },
  "schema_version": "1.0",
  "scoring": {
    "aggregate_definition": "arithmetic mean of the final per-case WebShop reward",
    "case_result_field": "result.result",
    "case_success": "result.result == 1.0",
    "failure_range": "0.0 <= result.result < 1.0",
    "primary_metric": "mean_score",
    "success_value": 1.0
  },
  "task_description": {
    "agent_role": "read the current page and available actions, then emit exactly one search or click tool selection",
    "objective": "find and purchase a product that satisfies the stated attributes and budget",
    "setting": "interactive text shopping website"
  },
  "test_split": "standard",
  "test_task": "webshop-std",
  "train_task": "webshop-env_train"
}
\end{lstlisting}

\begin{lstlisting}[
  style=prompttemplate,
  caption={Reasoning-backend information supplied during FH0 initialization.},
  label={lst:initializer-reasoning-backend}
]
{
  "backend_type": "ollama",
  "base_url": "http://127.0.0.1:11434",
  "credential_env": null,
  "generation": {
    "min_p": 0,
    "num_ctx": 32768,
    "num_predict": 4096,
    "presence_penalty": 1.5,
    "repeat_penalty": 1,
    "temperature": 0,
    "top_k": 20,
    "top_p": 0.95
  },
  "max_calls_per_step": 1,
  "max_concurrency": 1,
  "model_id": "Qwen3.8-27b",
  "retry_backoff_seconds": [
    1,
    2
  ],
  "retry_count": 2,
  "thinking": false,
  "timeout_seconds": 300
}
\end{lstlisting}

These forward-backend settings are distinct from the model controls of the
initializer process.  PLAN, BUILD, and REPAIR use fresh Codex CLI processes
configured with \texttt{gpt-5.6-sol}, high reasoning effort, and standard mode.
That coding model generates and repairs the harness, whereas the
Ollama/Qwen3.8-27b profile defines the inference backend called by the resulting
FH0 harness.

\subsection{Three-stage protocol}
\label{app:initializer-protocol}

\paragraph{PLAN.}
The controller publishes one immutable planning job for the complete network.
The planner reads only registered public inputs and produces a schema-valid,
function-level initialization plan.  For every node, the plan specifies its
  role, boundary behavior, prompt-level inductive bias, authorized tool exposure,
  error behavior, and acceptance checks.  It cannot inspect benchmark cases,
  modify an agent repository, alter the frozen topology, or request changes to
the fixed runtime mechanisms.

\paragraph{BUILD.}
After the plan is accepted, the controller materializes one repository per
node and publishes an independent build job for each node.  A builder may
specialize only \texttt{ROLE.md}, \texttt{prompts/system.md},
\texttt{src/harness\_agent/tools.py}, \texttt{tests/test\_smoke.py}, and the
allowlisted role metadata.  In particular, the fixed
\texttt{strategy.py}, \texttt{memory.py}, and \texttt{parsing.py} files remain
unchanged.  Each successful build runs the job-published checks from their exact
working directories.  The controller then
performs its own schema, repository, provenance, test, and contract validation.

\paragraph{REPAIR.}
If network validation reports a node-local failure, the controller publishes a
repair job with the structured failed checks, the frozen validation report, and
the corresponding probe requests and responses.  Repair preserves the original
node specification and may touch only the same builder-owned paths.  It may not
add dependencies, weaken tests, modify the fixed runtime, invent fallback
actions, or convert execution failures into successful outputs.  After each
repair round, the controller revalidates the network.  Initialization becomes
\texttt{ready} when all checks pass; after at most three repair rounds, remaining
failures produce a terminal \texttt{failed} result.

\subsection{Execution and acceptance separation}
\label{app:initializer-acceptance}

Every initializer job is executed by a fresh Codex CLI process with frozen
model controls.  The prompt templates specify \texttt{gpt-5.6-sol}, high
reasoning effort, and standard mode.  Model output is not itself an accepted
checkpoint.  The controller independently verifies the result envelope,
registered artifacts, allowed file changes, test results, and end-to-end
contract probes before accepting the node into FH0.  Raw execution traces and
controller-derived acceptance evidence are retained separately.

\subsection{PLAN prompt template}
\label{app:initializer-plan-prompt}

\begin{lstlisting}[
  style=prompttemplate,
  caption={Initializer PLAN prompt template.},
  label={lst:initializer-plan}
]
# HarnessNet initializer planner

Execute only when the immutable job manifest fixes `gpt-5.6-sol`, `reasoning_effort=high`, and `mode=standard`.
  Treat the manifest's registered read and write paths as the complete authority.

Design one function-level initialization plan for the supplied benchmark profile, backend profile, topology, stable
  harness contract, canonical template manifest, and plan schema. Do not read benchmark cases or modify an Agent
  repository. Describe every required inductive-bias field for every topology node. Persist only the schema-valid plan
  and its differentiation summary at the requested paths. For a single-layer topology, treat the sole node as both the
  first and last HarnessNet boundary: it consumes the current environment observation, replays the complete case-local
  history from case-local memory, and returns one exact admissible action in the adapter-accepted content form. The
  fixed runtime also appends audit records to the supplied memory path; use every ordered runtime record as
  interaction history with no truncation, no summary, and no separate harness-owned compact state. Preserve every
  native assistant message recorded by the Contract 1.2 template. When the public profile defines an adapter-owned
  `take_action` or other action tool, the final-boundary node must register the matching backend-native function tools
  through `tools.py` and the fixed SDK call. Non-final nodes must not register environment action tools. For every
  node, freeze the template's one-call native-message strategy. The plan must not introduce benchmark-solving
  heuristics, semantic repair, model retry, fallback actions, task-state summaries, or example-specific guidance. The
  builder may specialize only `ROLE.md`, `prompts/system.md`, `src/harness_agent/tools.py`, `tests/test_smoke.py`, and
  allowed role metadata. It must not modify `strategy.py`, `memory.py`, or `parsing.py`. Those plan fields must
  describe inherited template behavior, not request new implementations or stronger checks. Do not bypass these fixed
  files through tools.py or monkeypatching. The canonical parser preserves native assistant messages, structurally
  projects one native tool call to downstream content, and otherwise passes through nonempty native text. It rejects
  malformed tool-call structure and absent usable content; it does not validate benchmark tool-name membership or
  argument schemas. The benchmark adapter owns that action validation. Distinguish the model's instructed one-tool
  action format from harness-enforced rejection: register the exact public tool schemas and instruct the model to
  follow them, but do not require the builder to add unknown-tool, schema-argument, or nonempty-prose rejection to the
  harness. Keep acceptance checks within the fixed template's actual guarantees and the permitted role/tool
  specialization, including backend/upstream error propagation. Copy the frozen topology into the plan exactly,
  including its layer count, node placement, edges, and boundary IDs. A node serving more than one boundary role still
  appears only at its frozen position: never duplicate a node or add a layer to express those roles. In particular, a
  one-node topology remains one layer whose first and last Agent IDs are the same. For every node, keep model-output
  degradation separate from execution failure: `error_degradation_policy`, `prompt_guidance`, and `acceptance_checks`
  must require backend exceptions and upstream `status=error` inputs to propagate through the stable error envelope
  with `status=error` and a nonzero harness exit. They must never describe an `ok` fallback for those execution
  errors. Empty, malformed, or otherwise unusable successful model output must fail under Contract 1.2; FH0 must not
  invent content or return a degraded `status=ok` fallback. Copy `payload.plan_provenance` verbatim into the
  initialization plan's `provenance` object; never derive or modify any provenance field yourself. Also copy the
  frozen input refs exactly: set plan `benchmark_profile` to `payload.benchmark_profile_ref`, plan
  `llm_backend_profile` to `payload.llm_backend_profile_ref`, and plan `template_ref` to
  `payload.template_manifest_ref`. Do not replace these refs with profile summaries.

The public profile may describe topology, Agent interface, output/action schema, scoring semantics, SDK/tool access,
  backend controls, and the general task family. It must not expose or lead you to discover case paths, case IDs, case
  counts, case statistics, case observations, or case answers. Do not inspect conversation context or filesystem paths
  that are absent from the job manifest.

Run every `validation.checks` entry from its exact published `cwd`. These are job-agent checks only; controller-owned
  acceptance independently enforces schema, provenance, repository, pytest, checkpoint, and transaction contracts.

Use every path in `expected_result.artifact_paths` verbatim for its named result role, including each validation check
  ID; never substitute another writable path. Write the schema-valid result plus the required visible trace, tool
  events, stdout/stderr, usage, summaries, and artifact inventory only to those mapped output paths. If the job or
  model controls differ, write a schema-valid failed result instead of widening access or inventing inputs.
\end{lstlisting}

\subsection{BUILD prompt template}
\label{app:initializer-build-prompt}

\begin{lstlisting}[
  style=prompttemplate,
  caption={Initializer BUILD prompt template.},
  label={lst:initializer-build}
]
# HarnessNet initializer builder

Execute only when the immutable job manifest fixes `gpt-5.6-sol`, `reasoning_effort=high`, and `mode=standard`.
  Treat its registered read/write paths, validation commands, and result location as the complete authority.

Implement exactly one node specification in the materialized repository. Use the frozen initialization plan, fixed
  SDK/backend contract, and complete mutable path and manifest-field allowlists. Do not change identity, entrypoint,
  SDK/interface versions,
  dependencies, or runtime code. Preserve the stable execution-error contract exactly: backend exceptions and upstream
  `status=error` inputs must produce a stable `status=error` response and a nonzero harness exit; never catch them
  into a degraded `status=ok` response. Unusable output from an otherwise successful model call must fail under
  Contract 1.2; do not invent content or use degraded `ok` behavior. When the node uses the supplied case-local memory
  path, account for the fact that the fixed runtime also appends audit records there: preserve the complete case-local
  history with no truncation and no summary, including every recorded native assistant message. Configure
  backend-native function tools in `tools.py` only when this node owns the final action boundary, using the exact
  public-profile names and argument schemas. You may specialize only `ROLE.md`, `prompts/system.md`,
  `src/harness_agent/tools.py`, `tests/test_smoke.py`, and the allowed role metadata. You must not modify
  `strategy.py`; you must not modify `memory.py`; you must not modify `parsing.py`. Do not add benchmark-solving
  heuristics, compact task state, semantic repair, retries, or fallback actions. Include focused smoke tests for the
  node's role and backend-error propagation. Run every `validation.checks` entry from its exact published `cwd`. These
  are job-agent checks only; controller-owned acceptance independently enforces schema, provenance, repository,
  pytest, checkpoint, and transaction contracts. Use every path in `expected_result.artifact_paths` verbatim for its
  named result role, including each validation check ID; never substitute another writable path. Write the mapped
  repository diff and the schema-valid result, visible trace, tool events, stdout/stderr, usage, summary, and artifact
  inventory only at those mapped output paths. If any
  precondition fails, write a schema-valid failed result; do not repair or widen the job yourself.
\end{lstlisting}

\subsection{REPAIR prompt template}
\label{app:initializer-repair-prompt}

\begin{lstlisting}[
  style=prompttemplate,
  caption={Initializer REPAIR prompt template.},
  label={lst:initializer-repair}
]
# HarnessNet initializer repair

Execute only when the immutable job manifest fixes `gpt-5.6-sol`, `reasoning_effort=high`, and `mode=standard`.
  Require `repair_limit=3` and a `repair_round` of 1, 2, or 3; reject any job outside that frozen limit.

Repair only the structured failed checks for the supplied node. Before changing the repository,
  read `payload.benchmark_profile_ref`, the frozen `payload.validation_report_ref`, and every artifact in
  `payload.probe_diagnostic_refs`. Treat the registered probe requests and responses as the exact observed wire
  evidence for this repair; compare them with the public benchmark profile instead of guessing from a check ID or
  summary message alone. Keep the original node specification and network inductive biases fixed. Do not add
  dependencies or weaken tests. Repair may change only builder-owned `ROLE.md`, `prompts/system.md`,
  `src/harness_agent/tools.py`, role metadata, and focused smoke tests. It must not modify the fixed Contract 1.2
  `strategy.py`, `memory.py`, or `parsing.py` mechanisms. A failure located in those fixed mechanisms is out of scope
  for per-node repair. Never convert backend exceptions or upstream `status=error` inputs into a degraded `status=ok`
  response: the stable contract requires an error envelope and nonzero harness exit for execution errors. Do not
  convert unusable model output into invented content or a degraded successful fallback. Use the prior repair record
  when supplied and run every `validation.checks` entry from its exact published `cwd`. These are job-agent checks
  only; controller-owned acceptance independently enforces schema, provenance, repository, pytest, checkpoint, and
  transaction contracts. Then use every path in `expected_result.artifact_paths` verbatim for its named result role,
  including each validation check ID; never substitute another writable path. Write the mapped repository diff and
  the schema-valid result, visible trace, tool events, stdout/stderr, usage, summary, and
  artifact inventory only at those mapped paths. If the supplied failure cannot be repaired within the allowlist
  because the frozen probe evidence locates it in another node, do not modify the repository. Write a schema-valid
  `status=failed`, `outcome=unchanged` result with no repository changes, empty `addressed_check_ids`, the original
  failed IDs in `remaining_failed_check_ids`, and one non-retriable
  `repair_out_of_scope` error. The controller may accept only that exact no-change shape so the network can be
  revalidated after the responsible nodes finish.
\end{lstlisting}

\subsection{Canonical harness template}
\label{app:initializer-harness-template}

Every node repository starts from the same canonical Contract~1.2 harness
template.  Listing~\ref{lst:harness-template-tree} shows the complete template
tree before node identity is materialized.  The controller subsequently fills
the topology-derived identity fields in \texttt{harness.json}; it does not add
benchmark instances or node-specific solving logic.

\begin{lstlisting}[
  caption={File structure of the canonical per-node harness template.},
  label={lst:harness-template-tree}
]
agent_repo/
|-- harness.json
|-- pyproject.toml
|-- README.md
|-- ROLE.md
|-- prompts/
|   |-- forward.md
|   `-- system.md
|-- src/
|   `-- harness_agent/
|       |-- __init__.py
|       |-- entrypoint.py
|       |-- memory.py
|       |-- parsing.py
|       |-- strategy.py
|       `-- tools.py
`-- tests/
    `-- test_smoke.py
\end{lstlisting}

Table~\ref{tab:harness-template-files} summarizes the responsibility and
initializer-time ownership of every file.  Here, ``fixed'' means fixed during
FH0 initialization; it does not redefine the separate optimizer-time
allowlist.

\begin{table}[H]
  \centering
  \caption{Files in the canonical harness template and their initializer-time roles.}
  \label{tab:harness-template-files}
  \scriptsize
  \setlength{\tabcolsep}{4pt}
  \renewcommand{\arraystretch}{1.12}
  \begin{tabularx}{\textwidth}{@{}p{0.29\textwidth}p{0.18\textwidth}X@{}}
    \toprule
    \textbf{File} & \textbf{Initializer treatment} & \textbf{Purpose} \\
    \midrule
    \texttt{harness.json} & Controller-owned; scoped metadata editable &
    Declares node identity, topology neighbors, entry point, contract versions,
    template metadata, and path/field allowlists.  The materializer writes the
    identity fields; the builder may change only the allowed role metadata. \\
    \texttt{pyproject.toml} & Fixed &
    Defines the self-contained Python package, supported interpreter version,
    dependencies, and test discovery settings. \\
    \texttt{README.md} & Fixed &
    Records the neutral Contract~1.2 baseline, invocation command, and editing
    boundary for the repository. \\
    \texttt{ROLE.md} & Builder-owned &
    Provides the auditable, node-specific role description derived from the
    accepted initialization plan. \\
    \texttt{prompts/system.md} & Builder-owned &
    Supplies the node's runtime system instruction and its plan-derived
    inductive bias. \\
    \texttt{prompts/forward.md} & Fixed &
    Retains a future trainable prompt extension point; the neutral FH0 strategy
    does not inject this file into the model request. \\
    \texttt{src/harness\_agent/}\newline
    \texttt{\_\_init\_\_.py} & Fixed &
    Marks the executable harness package and states its neutral baseline role. \\
    \texttt{src/harness\_agent/}\newline
    \texttt{entrypoint.py} & Fixed &
    Implements the subprocess request/response boundary, constructs the SDK
    client from the supplied backend profile, and preserves the stable error
    exit contract. \\
    \texttt{src/harness\_agent/}\newline
    \texttt{memory.py} & Fixed mechanism &
    Loads the complete case-local interaction history in original order without
    truncation or summarization. \\
    \texttt{src/harness\_agent/}\newline
    \texttt{parsing.py} & Fixed mechanism &
    Projects native assistant content or one native tool call into the stable
    downstream envelope and rejects unusable model output. \\
    \texttt{src/harness\_agent/}\newline
    \texttt{strategy.py} & Fixed mechanism &
    Reconstructs messages from the full local history, appends current upstream
    inputs, makes one backend call, and delegates output projection. \\
    \texttt{src/harness\_agent/}\newline
    \texttt{tools.py} & Builder-owned &
    Registers backend-native function schemas only when the node owns the public
    benchmark action boundary. \\
    \texttt{tests/test\_smoke.py} & Builder-owned &
    Contains focused node-role, output-contract, and error-propagation checks
    executed before controller acceptance. \\
    \bottomrule
  \end{tabularx}
\end{table}

The template enters the initializer workflow at five points.  First, the
controller freezes the clean \texttt{template-baseline} and its manifest.
Second, PLAN reads that public template contract and proposes only node-specific
roles, prompt biases, tools, and acceptance checks; it does not edit a
repository.  Third, after plan acceptance, the deterministic materializer
creates an independent Git repository for each node and fills its topology
identity in \texttt{harness.json}.  Fourth, BUILD specializes only
\texttt{ROLE.md}, \texttt{prompts/system.md},
\texttt{src/harness\_agent/tools.py}, \texttt{tests/test\_smoke.py}, and the
allowlisted role metadata.  Finally, the controller checks required files,
fixed mechanisms, manifest fields, repository cleanliness, imports, tests, and
synthetic network probes.  REPAIR receives the same template-derived boundary
and may correct only builder-owned files.  The accepted node repositories
jointly constitute FH0.

\clearpage
\section{Forward-Model Configurations}
\label{app:model-configurations}

Table~\ref{tab:ollama-model-controls} reports the frozen model controls used
across the experiments.  Local registry identifiers are omitted.  Parameters
not listed in the frozen generation objects use the corresponding model and
Ollama defaults.

\begin{table}[H]
  \centering
  \caption{Ollama model and execution controls used in the HarnessNet
  experiments.  The Qwen3.8-27b context and prediction limits are the only
  generation-control differences among the four configurations.}
  \label{tab:ollama-model-controls}
  \scriptsize
  \setlength{\tabcolsep}{3.2pt}
  \renewcommand{\arraystretch}{1.12}
  \begin{tabularx}{\textwidth}{@{}l*{4}{>{\centering\arraybackslash}X}@{}}
    \toprule
    \textbf{Parameter} &
    \textbf{Qwen3.8-27b} &
    \textbf{Qwen3.5-9b} &
    \textbf{Gemma4-8b-e4b} &
    \textbf{Ornith-1.5-9B} \\
    \midrule
    Resource &
    \href{https://huggingface.co/unsloth/Qwen3.8-27B-GGUF}{Hugging Face} &
    \href{https://huggingface.co/unsloth/Qwen3.5-9B-GGUF}{Hugging Face} &
    \href{https://huggingface.co/unsloth/gemma-4-E4B-it-MLX-8bit}{Hugging Face} &
    \href{https://huggingface.co/ornith-ai/Ornith-1.5-9B-GGUF}{Hugging Face} \\
    Quantization & Q4 & Q4 & Q8 & Q4 \\
    Backend & Ollama & Ollama & Ollama & Ollama \\
    \midrule
    \texttt{num\_ctx} & 32,768 & 20,480 & 20,480 & 20,480 \\
    \texttt{num\_predict} & 4,096 & 2,048 & 2,048 & 2,048 \\
    \texttt{temperature} & 0 & 0 & 0 & 0 \\
    \texttt{top\_p} & 0.95 & 0.95 & 0.95 & 0.95 \\
    \texttt{top\_k} & 20 & 20 & 20 & 20 \\
    \texttt{min\_p} & 0 & 0 & 0 & 0 \\
    \texttt{presence\_penalty} & 1.5 & 1.5 & 1.5 & 1.5 \\
    \texttt{repeat\_penalty} & 1 & 1 & 1 & 1 \\
    Thinking & false & false & false & false \\
    \midrule
    Max calls per step & 1 & 1 & 1 & 1 \\
    Max concurrency & 1 & 1 & 1 & 1 \\
    Timeout (s) & 300 & 300 & 300 & 300 \\
    Maximum timeout retries & 2 & 2 & 2 & 2 \\
    Retry backoff (s) & 1, 2 & 1, 2 & 1, 2 & 1, 2 \\
    \bottomrule
  \end{tabularx}
\end{table}

\clearpage
\section{Training Experiments}
\label{app:training-experiments}

\begingroup
\setlength{\intextsep}{6pt plus 1pt minus 1pt}
\captionsetup[table]{skip=3pt}
\renewcommand{\arraystretch}{0.88}

This appendix reports the training curves and held-out test results for the
three evaluated topologies on ALFWorld and WebShop.  For ALFWorld, scores are
pass rates; for WebShop, scores are mean final rewards (AvgScore).  An
asterisk in the \emph{Selected} column marks the trained checkpoint chosen
for selected-FH testing.  The \emph{Model-only} row denotes the harness-free
baseline.  Held-out test scores are shown on the FH0 row and on the selected
checkpoint row; the remaining checkpoint test cells are intentionally blank.

\noindent\textbf{Subsection navigation}
\par\smallskip
\begingroup
\small
\setlength{\tabcolsep}{8pt}
\renewcommand{\arraystretch}{1.08}
\noindent\begin{tabularx}{\textwidth}{@{}>{\raggedright\arraybackslash}X@{}}
  \toprule
  \hyperref[app:training-alfworld-agentron]{\ref*{app:training-alfworld-agentron}\enspace ALFWorld-Agentron-Training} \\
  \hyperref[app:training-webshop-agentron]{\ref*{app:training-webshop-agentron}\enspace WebShop-Agentron-Training} \\
  \hyperref[app:training-alfworld-2node]{\ref*{app:training-alfworld-2node}\enspace ALFWorld-2Node HarnessNet-Training} \\
  \hyperref[app:training-webshop-2node]{\ref*{app:training-webshop-2node}\enspace WebShop-2Node HarnessNet-Training} \\
  \hyperref[app:training-alfworld-4node]{\ref*{app:training-alfworld-4node}\enspace ALFWorld-4Node HarnessNet-Training} \\
  \hyperref[app:training-webshop-4node]{\ref*{app:training-webshop-4node}\enspace WebShop-4Node HarnessNet-Training} \\
  \bottomrule
\end{tabularx}
\endgroup
\medskip

\subsection{ALFWorld-Agentron-Training}
\label{app:training-alfworld-agentron}

\begin{table}[H]
  \centering
  \caption{Key metadata for ALFWorld-Agentron training.}
  \label{tab:alfworld-agentron-metadata}
  \footnotesize
  \begin{tabular}{@{}ll@{}}
    \toprule
    \textbf{Field} & \textbf{Value} \\
    \midrule
    Forward model & Qwen3.8-27b \\
    Benchmark domain & ALFWorld \\
    Topology & Agentron $[1]$ \\
    Train dataset & \texttt{alfworld-env\_train} ($n=168$) \\
    Validation dataset & \texttt{alfworld-env\_train} ($n=42$) \\
    Test dataset & \texttt{alfworld-std}, \texttt{new\_std} split ($n=109$) \\
    \bottomrule
  \end{tabular}
\end{table}

\begin{table}[H]
  \centering
  \caption{ALFWorld-Agentron train, validation, and held-out test records.}
  \label{tab:alfworld-agentron-results}
  \footnotesize
  \begin{tabular}{@{}ccccc@{}}
    \toprule
    \textbf{Checkpoint} & \textbf{\shortstack{Train\\Score}} & \textbf{\shortstack{Validation\\Score}} & \textbf{Selected} & \textbf{\shortstack{Test\\Score}} \\
    \midrule
    Model-only & & & & 56.88\% \\
    FH0 & 59.18\% & 40.48\% & & 22.02\% \\
    FH1 & 39.39\% & 52.38\% & & \\
    FH2 & 55.56\% & 66.67\% & & \\
    FH3 & 58.54\% & 66.67\% & & \\
    FH4 & 76.47\% & 83.33\% & & \\
    FH5 & 83.13\% & 85.71\% & & \\
    FH6 & 87.58\% & 80.95\% & * & 77.98\% \\
    FH7 & 88.10\% & 78.57\% & & \\
    \bottomrule
  \end{tabular}
\end{table}

\subsection{WebShop-Agentron-Training}
\label{app:training-webshop-agentron}

\begin{table}[H]
  \centering
  \caption{Key metadata for WebShop-Agentron training.}
  \label{tab:webshop-agentron-metadata}
  \footnotesize
  \begin{tabular}{@{}ll@{}}
    \toprule
    \textbf{Field} & \textbf{Value} \\
    \midrule
    Forward model & Qwen3.8-27b \\
    Benchmark domain & WebShop \\
    Topology & Agentron $[1]$ \\
    Train dataset & \texttt{webshop-env\_train} ($n=320$) \\
    Validation dataset & \texttt{webshop-env\_train} ($n=80$) \\
    Test dataset & \texttt{webshop-std}, \texttt{standard} split ($n=200$) \\
    \bottomrule
  \end{tabular}
\end{table}

\begin{table}[H]
  \centering
  \caption{WebShop-Agentron train, validation, and held-out test records.}
  \label{tab:webshop-agentron-results}
  \footnotesize
  \begin{tabular}{@{}ccccc@{}}
    \toprule
    \textbf{Checkpoint} & \textbf{\shortstack{Train\\Score}} & \textbf{\shortstack{Validation\\Score}} & \textbf{Selected} & \textbf{\shortstack{Test\\Score}} \\
    \midrule
    Model-only & & & & 0.5670 \\
    FH0 & 0.4789 & 0.5357 & & 0.5559 \\
    FH1 & 0.5047 & 0.6384 & & \\
    FH2 & 0.6691 & 0.6615 & & \\
    FH3 & 0.7078 & 0.7223 & & \\
    FH4 & 0.6863 & 0.7010 & & \\
    FH5 & 0.7250 & 0.6583 & & \\
    FH6 & 0.6397 & 0.6935 & & \\
    FH7 & 0.7087 & 0.6867 & & \\
    FH8 & 0.6699 & 0.7075 & & \\
    FH9 & 0.7153 & 0.7131 & & \\
    FH10 & 0.7115 & 0.7222 & * & 0.6950 \\
    \bottomrule
  \end{tabular}
\end{table}

\subsection{ALFWorld-2Node HarnessNet-Training}
\label{app:training-alfworld-2node}

\begin{table}[H]
  \centering
  \caption{Key metadata for ALFWorld-2Node HarnessNet training.}
  \label{tab:alfworld-2node-metadata}
  \footnotesize
  \begin{tabular}{@{}ll@{}}
    \toprule
    \textbf{Field} & \textbf{Value} \\
    \midrule
    Forward model & Qwen3.8-27b \\
    Benchmark domain & ALFWorld \\
    Topology & 2Node HarnessNet $[1,1]$ \\
    Train dataset & \texttt{alfworld-env\_train} ($n=168$) \\
    Validation dataset & \texttt{alfworld-env\_train} ($n=42$) \\
    Test dataset & \texttt{alfworld-std}, \texttt{new\_std} split ($n=109$) \\
    \bottomrule
  \end{tabular}
\end{table}

\begin{table}[H]
  \centering
  \caption{ALFWorld-2Node HarnessNet train, validation, and held-out test records.}
  \label{tab:alfworld-2node-results}
  \footnotesize
  \begin{tabular}{@{}ccccc@{}}
    \toprule
    \textbf{Checkpoint} & \textbf{\shortstack{Train\\Score}} & \textbf{\shortstack{Validation\\Score}} & \textbf{Selected} & \textbf{\shortstack{Test\\Score}} \\
    \midrule
    Model-only & & & & 56.88\% \\
    FH0 & 44.44\% & 57.14\% & & 62.39\% \\
    FH1 & 60.42\% & 71.43\% & & \\
    FH2 & 45.21\% & 47.62\% & & \\
    FH3 & 89.47\% & 85.71\% & & \\
    FH4 & 91.67\% & 85.71\% & & \\
    FH5 & 92.26\% & 85.71\% & * & 85.32\% \\
    \bottomrule
  \end{tabular}
\end{table}

\subsection{WebShop-2Node HarnessNet-Training}
\label{app:training-webshop-2node}

\begin{table}[H]
  \centering
  \caption{Key metadata for WebShop-2Node HarnessNet training.}
  \label{tab:webshop-2node-metadata}
  \footnotesize
  \begin{tabular}{@{}ll@{}}
    \toprule
    \textbf{Field} & \textbf{Value} \\
    \midrule
    Forward model & Qwen3.8-27b \\
    Benchmark domain & WebShop \\
    Topology & 2Node HarnessNet $[1,1]$ \\
    Train dataset & \texttt{webshop-env\_train} ($n=320$) \\
    Validation dataset & \texttt{webshop-env\_train} ($n=80$) \\
    Test dataset & \texttt{webshop-std}, \texttt{standard} split ($n=200$) \\
    \bottomrule
  \end{tabular}
\end{table}

\begin{table}[H]
  \centering
  \caption{WebShop-2Node HarnessNet train, validation, and held-out test records.}
  \label{tab:webshop-2node-results}
  \footnotesize
  \begin{tabular}{@{}ccccc@{}}
    \toprule
    \textbf{Checkpoint} & \textbf{\shortstack{Train\\Score}} & \textbf{\shortstack{Validation\\Score}} & \textbf{Selected} & \textbf{\shortstack{Test\\Score}} \\
    \midrule
    Model-only & & & & 0.5670 \\
    FH0 & 0.4807 & 0.5852 & & 0.5523 \\
    FH1 & 0.6021 & 0.5998 & & \\
    FH2 & 0.5814 & 0.6450 & & \\
    FH3 & 0.5429 & 0.5885 & & \\
    FH4 & 0.6282 & 0.6054 & & \\
    FH5 & 0.6778 & 0.6094 & & \\
    FH6 & 0.6437 & 0.6163 & * & 0.5892 \\
    FH7 & 0.5755 & 0.6444 & & \\
    \bottomrule
  \end{tabular}
\end{table}

\subsection{ALFWorld-4Node HarnessNet-Training}
\label{app:training-alfworld-4node}

\begin{table}[H]
  \centering
  \caption{Key metadata for ALFWorld-4Node HarnessNet training.}
  \label{tab:alfworld-4node-metadata}
  \footnotesize
  \begin{tabular}{@{}ll@{}}
    \toprule
    \textbf{Field} & \textbf{Value} \\
    \midrule
    Forward model & Qwen3.5-9b \\
    Benchmark domain & ALFWorld \\
    Topology & 4Node HarnessNet $[1,2,1]$ \\
    Train dataset & \texttt{alfworld-env\_train} ($n=168$) \\
    Validation dataset & \texttt{alfworld-env\_train} ($n=42$) \\
    Test dataset & \texttt{alfworld-std}, \texttt{new\_std} split ($n=109$) \\
    \bottomrule
  \end{tabular}
\end{table}

\begin{table}[H]
  \centering
  \caption{ALFWorld-4Node HarnessNet train, validation, and held-out test records.}
  \label{tab:alfworld-4node-results}
  \footnotesize
  \begin{tabular}{@{}ccccc@{}}
    \toprule
    \textbf{Checkpoint} & \textbf{\shortstack{Train\\Score}} & \textbf{\shortstack{Validation\\Score}} & \textbf{Selected} & \textbf{\shortstack{Test\\Score}} \\
    \midrule
    Model-only & & & & 11.93\% \\
    FH0 & 9.09\% & 19.05\% & & 16.51\% \\
    FH1 & 34.43\% & 33.33\% & & \\
    FH2 & 33.33\% & 28.57\% & & \\
    FH3 & 39.39\% & 38.10\% & & \\
    FH4 & 42.86\% & 38.10\% & & \\
    FH5 & 79.17\% & 76.19\% & & \\
    FH6 & 79.17\% & 83.33\% & * & 69.72\% \\
    FH7 & 85.71\% & 83.33\% & & \\
    \bottomrule
  \end{tabular}
\end{table}

\subsection{WebShop-4Node HarnessNet-Training}
\label{app:training-webshop-4node}

\begin{table}[H]
  \centering
  \caption{Key metadata for WebShop-4Node HarnessNet training.}
  \label{tab:webshop-4node-metadata}
  \footnotesize
  \begin{tabular}{@{}ll@{}}
    \toprule
    \textbf{Field} & \textbf{Value} \\
    \midrule
    Forward model & Gemma4-8b-e4b \\
    Benchmark domain & WebShop \\
    Topology & 4Node HarnessNet $[1,2,1]$ \\
    Train dataset & \texttt{webshop-env\_train} ($n=320$) \\
    Validation dataset & \texttt{webshop-env\_train} ($n=80$) \\
    Test dataset & \texttt{webshop-std}, \texttt{standard} split ($n=200$) \\
    \bottomrule
  \end{tabular}
\end{table}

\begin{table}[H]
  \centering
  \caption{WebShop-4Node HarnessNet train, validation, and held-out test records.}
  \label{tab:webshop-4node-results}
  \footnotesize
  \begin{tabular}{@{}ccccc@{}}
    \toprule
    \textbf{Checkpoint} & \textbf{\shortstack{Train\\Score}} & \textbf{\shortstack{Validation\\Score}} & \textbf{Selected} & \textbf{\shortstack{Test\\Score}} \\
    \midrule
    Model-only & & & & 0.4083 \\
    FH0 & 0.1210 & 0.0963 & & 0.0624 \\
    FH1 & 0.4755 & 0.5298 & & \\
    FH2 & 0.5984 & 0.6167 & * & 0.5875 \\
    FH3 & 0.4837 & 0.5379 & & \\
    FH4 & 0.5710 & 0.4727 & & \\
    FH5 & 0.6172 & 0.5408 & & \\
    FH6 & 0.6085 & 0.5509 & & \\
    FH7 & 0.5597 & 0.5435 & & \\
    \bottomrule
  \end{tabular}
\end{table}

\endgroup

\clearpage
\section{Migration Study}
\label{app:migration-study}

\setlength{\intextsep}{6pt plus 1pt minus 1pt}
\captionsetup[table]{skip=3pt}

The migration study evaluates whether the forward-harness checkpoints learned
with one node-level language model remain useful when the underlying model is
replaced.  Table~\ref{tab:migration-metadata} summarizes the common evaluation
setting.

\begin{table}[H]
  \centering
  \caption{Key metadata for the migration study.}
  \label{tab:migration-metadata}
  \small
  \begin{tabular}{@{}ll@{}}
    \toprule
    \textbf{Field} & \textbf{Value} \\
    \midrule
    Forward models & Qwen3.5-9B; Ornith1.5-9B; gemma4-8B-E4B \\
    Benchmark domain & ALFWorld \\
    Topology & 4Node HarnessNet $[1,2,1]$ \\
    Validation dataset & \texttt{alfworld-env\_train} ($n=42$) \\
    Test dataset & \texttt{alfworld-std}, \texttt{new\_std} split ($n=109$) \\
    \bottomrule
  \end{tabular}
\end{table}

\begin{table}[H]
  \centering
  \caption{Migration-study validation and held-out test pass rates.}
  \label{tab:migration-results}
  \small
  \setlength{\tabcolsep}{10pt}
  \renewcommand{\arraystretch}{1.00}
  \begin{tabular}{@{}lccc@{}}
    \toprule
    \textbf{Checkpoint/Model} &
    \textbf{Qwen3.5-9B} &
    \textbf{Ornith1.5-9B} &
    \textbf{gemma4-8B-E4B} \\
    \midrule
    FH0 & 19.05\% & 16.67\% & 16.67\% \\
    FH1 & 33.33\% & 19.05\% & 16.67\% \\
    FH2 & 28.57\% & 23.81\% & 7.14\% \\
    FH3 & 38.10\% & 28.57\% & 30.95\% \\
    FH4 & 38.10\% & 33.33\% & 38.10\% \\
    FH5 & 76.19\% & 57.14\% & 78.57\% \\
    FH6 & 83.33\% & 61.90\% & 73.81\% \\
    FH7 & 83.33\% & 59.52\% & 80.95\% \\
    \midrule
    Model-only Test & 11.93\% & 19.27\% & 8.26\% \\
    FH0 Test & 16.51\% & 11.93\% & 9.17\% \\
    Selected FH & FH6 & FH6 & FH7 \\
    Selected FH Test & 69.72\% & 49.54\% & 68.81\% \\
    \bottomrule
  \end{tabular}
\end{table}

Across all three model backends, later harness checkpoints substantially
outperform FH0 on validation: the selected checkpoint reaches 83.33\% for
Qwen3.5-9B, 61.90\% for Ornith1.5-9B, and 80.95\% for gemma4-8B-E4B, compared
with FH0 scores of 19.05\%, 16.67\%, and 16.67\%, respectively.  The selected
  checkpoints also retain nontrivial held-out test pass rates after migration
(69.72\%, 49.54\%, and 68.81\%).  The node harnesses therefore preserve useful
task-solving structure after the underlying model is replaced.  Their learned
mechanisms and rules are not tied exclusively to the model used during training.
The variation across models shows that this transfer is partial rather than
fully model-independent.

\clearpage
\section{Hierarchical vs. Global Optimizer Comparison Study Details}
\label{app:optimizer-ablation}

\begingroup
\setlength{\intextsep}{6pt plus 1pt minus 1pt}
\captionsetup[table]{skip=3pt}
\renewcommand{\arraystretch}{0.92}

\noindent\textbf{Subsection navigation}
\par\smallskip
\begingroup
\small
\setlength{\tabcolsep}{8pt}
\renewcommand{\arraystretch}{1.08}
\noindent\begin{tabularx}{\textwidth}{@{}>{\raggedright\arraybackslash}X@{}}
  \toprule
  \hyperref[app:optimizer-ablation-results]{\ref*{app:optimizer-ablation-results}\enspace Training and Evaluation Results} \\
  \hyperref[app:optimizer-ablation-cost]{\ref*{app:optimizer-ablation-cost}\enspace Optimizer Cost Comparison} \\
  \hyperref[app:optimizer-ablation-prompts]{\ref*{app:optimizer-ablation-prompts}\enspace Optimizer Prompt Templates} \\
  \bottomrule
\end{tabularx}
\endgroup
\medskip

\subsection{Training and Evaluation Results}
\label{app:optimizer-ablation-results}

Tables~\ref{tab:optimizer-ablation-alfworld} and
\ref{tab:optimizer-ablation-webshop} compare checkpoint-level performance when
the same 4Node HarnessNet topology is trained with either the hierarchical
optimizer or the global optimizer.  ALFWorld scores are pass rates, whereas
WebShop scores are mean final rewards (AvgScore).  \emph{Model-only} denotes
the harness-free baseline.  The FH0 row combines the untrained harness's train
and validation scores with its no-training held-out test score.  Blank test
cells indicate that no test result was recorded for that checkpoint.

\begin{table}[H]
  \centering
  \caption{ALFWorld train, validation, and held-out test pass rates under the
  hierarchical and global optimizers.}
  \label{tab:optimizer-ablation-alfworld}
  \scriptsize
  \setlength{\tabcolsep}{3.2pt}
  \begin{tabular}{@{}lcccccc@{}}
    \toprule
    & \multicolumn{3}{c}{\textbf{Hierarchical Optimizer}}
    & \multicolumn{3}{c}{\textbf{Global Optimizer}} \\
    \cmidrule(lr){2-4}\cmidrule(l){5-7}
    \textbf{Checkpoint}
    & \textbf{\shortstack{Train\\Score}}
    & \textbf{\shortstack{Validation\\Score}}
    & \textbf{\shortstack{Test\\Score}}
    & \textbf{\shortstack{Train\\Score}}
    & \textbf{\shortstack{Validation\\Score}}
    & \textbf{\shortstack{Test\\Score}} \\
    \midrule
    Model-only & & & 11.93\% & & & 11.93\% \\
    FH0 & 9.09\% & 19.05\% & 16.51\% & 9.09\% & 19.05\% & 16.51\% \\
    FH1 & 34.43\% & 33.33\% & & 32.20\% & 28.57\% & \\
    FH2 & 33.33\% & 28.57\% & & 23.08\% & 30.95\% & \\
    FH3 & 39.39\% & 38.10\% & & 32.20\% & 45.24\% & \\
    FH4 & 42.86\% & 38.10\% & & 34.43\% & 40.48\% & \\
    FH5 & 79.17\% & 76.19\% & & 35.48\% & 59.52\% & \\
    FH6 & 79.17\% & 83.33\% & 69.72\% & 23.08\% & 38.10\% & \\
    FH7 & 85.71\% & 83.33\% & & 47.37\% & 42.86\% & 44.96\% \\
    \bottomrule
  \end{tabular}
\end{table}

\begin{table}[H]
  \centering
  \caption{WebShop train, validation, and held-out test AvgScore under the
  hierarchical and global optimizers.}
  \label{tab:optimizer-ablation-webshop}
  \scriptsize
  \setlength{\tabcolsep}{3.2pt}
  \begin{tabular}{@{}lcccccc@{}}
    \toprule
    & \multicolumn{3}{c}{\textbf{Hierarchical Optimizer}}
    & \multicolumn{3}{c}{\textbf{Global Optimizer}} \\
    \cmidrule(lr){2-4}\cmidrule(l){5-7}
    \textbf{Checkpoint}
    & \textbf{\shortstack{Train\\Score}}
    & \textbf{\shortstack{Validation\\Score}}
    & \textbf{\shortstack{Test\\Score}}
    & \textbf{\shortstack{Train\\Score}}
    & \textbf{\shortstack{Validation\\Score}}
    & \textbf{\shortstack{Test\\Score}} \\
    \midrule
    Model-only & & & 0.4083 & & & 0.4083 \\
    FH0 & 0.1210 & 0.0963 & 0.0624 & 0.1210 & 0.0963 & 0.0624 \\
    FH1 & 0.4755 & 0.5298 & & 0.3883 & 0.3802 & \\
    FH2 & 0.5984 & 0.6167 & 0.5875 & 0.4811 & 0.4730 & \\
    FH3 & 0.4837 & 0.5379 & & 0.4837 & 0.5554 & \\
    FH4 & 0.5710 & 0.4727 & & 0.6273 & 0.6418 & 0.6046 \\
    FH5 & 0.6172 & 0.5408 & & 0.6017 & 0.5608 & \\
    FH6 & 0.6085 & 0.5509 & & 0.6128 & 0.5894 & \\
    FH7 & 0.5597 & 0.5435 & & 0.5739 & 0.5747 & \\
    \bottomrule
  \end{tabular}
\end{table}

\subsection{Optimizer Cost Comparison}
\label{app:optimizer-ablation-cost}

The cost tables report optimizer usage for each backward round.  Costs for the
hierarchical optimizer are listed separately for the four node-level jobs.  The
global optimizer values span those four rows because they correspond to one
joint network-level job.  Optimizer time is converted from milliseconds to
seconds, and all reported values are rounded to the nearest integer.

\begin{table}[H]
  \centering
  \caption{ALFWorld optimizer cost by backward round.}
  \label{tab:optimizer-cost-alfworld}
  \scriptsize
  \setlength{\tabcolsep}{1.9pt}
  \renewcommand{\arraystretch}{0.86}
  \begin{tabular}{@{}clrrrrrr@{}}
    \toprule
    & \multicolumn{4}{c}{\textbf{Hierarchical Optimizer}}
    & \multicolumn{3}{c}{\textbf{Global Optimizer}} \\
    \cmidrule(lr){2-5}\cmidrule(l){6-8}
    \textbf{Backward}
    & \textbf{Node}
    & \textbf{\shortstack{Input\\Token}}
    & \textbf{\shortstack{Output\\Token}}
    & \textbf{\shortstack{Optimizer\\Time (s)}}
    & \textbf{\shortstack{Input\\Token}}
    & \textbf{\shortstack{Output\\Token}}
    & \textbf{\shortstack{Optimizer\\Time (s)}} \\
    \midrule
    \multirow{4}{*}{1} & \texttt{l2-last} & 4,557,381 & 31,320 & 1,427 & \multirow{4}{*}{10,467,460} & \multirow{4}{*}{40,902} & \multirow{4}{*}{1,719} \\
    & \texttt{l1-hidden\_a} & 4,440,628 & 24,540 & 1,304 & & & \\
    & \texttt{l1-hidden\_b} & 6,460,733 & 25,899 & 1,366 & & & \\
    & \texttt{l0-first} & 7,133,103 & 25,820 & 1,467 & & & \\
    \cmidrule(lr){2-8}
    \multirow{4}{*}{2} & \texttt{l2-last} & 7,027,571 & 46,165 & 1,912 & \multirow{4}{*}{13,542,625} & \multirow{4}{*}{42,660} & \multirow{4}{*}{2,056} \\
    & \texttt{l1-hidden\_a} & 11,152,144 & 43,697 & 1,779 & & & \\
    & \texttt{l1-hidden\_b} & 5,910,337 & 31,320 & 1,414 & & & \\
    & \texttt{l0-first} & 10,529,967 & 39,392 & 1,970 & & & \\
    \cmidrule(lr){2-8}
    \multirow{4}{*}{3} & \texttt{l2-last} & 5,084,140 & 32,673 & 1,355 & \multirow{4}{*}{13,214,124} & \multirow{4}{*}{71,496} & \multirow{4}{*}{2,290} \\
    & \texttt{l1-hidden\_a} & 5,374,440 & 26,883 & 1,292 & & & \\
    & \texttt{l1-hidden\_b} & 7,538,927 & 31,301 & 1,712 & & & \\
    & \texttt{l0-first} & 4,620,862 & 33,177 & 1,374 & & & \\
    \cmidrule(lr){2-8}
    \multirow{4}{*}{4} & \texttt{l2-last} & 3,348,882 & 23,800 & 940 & \multirow{4}{*}{8,173,538} & \multirow{4}{*}{48,924} & \multirow{4}{*}{1,452} \\
    & \texttt{l1-hidden\_a} & 4,835,618 & 28,256 & 1,123 & & & \\
    & \texttt{l1-hidden\_b} & 4,043,002 & 17,644 & 885 & & & \\
    & \texttt{l0-first} & 3,967,817 & 18,529 & 774 & & & \\
    \cmidrule(lr){2-8}
    \multirow{4}{*}{5} & \texttt{l2-last} & 2,288,059 & 14,149 & 576 & \multirow{4}{*}{12,491,469} & \multirow{4}{*}{43,673} & \multirow{4}{*}{1,873} \\
    & \texttt{l1-hidden\_a} & 4,746,387 & 27,297 & 1,159 & & & \\
    & \texttt{l1-hidden\_b} & 3,398,826 & 18,563 & 816 & & & \\
    & \texttt{l0-first} & 3,739,480 & 23,164 & 959 & & & \\
    \cmidrule(lr){2-8}
    \multirow{4}{*}{6} & \texttt{l2-last} & 3,641,378 & 17,629 & 729 & \multirow{4}{*}{7,927,776} & \multirow{4}{*}{44,207} & \multirow{4}{*}{1,436} \\
    & \texttt{l1-hidden\_a} & 4,242,029 & 30,355 & 1,213 & & & \\
    & \texttt{l1-hidden\_b} & 3,211,536 & 17,740 & 755 & & & \\
    & \texttt{l0-first} & 3,144,400 & 24,552 & 1,006 & & & \\
    \cmidrule(lr){2-8}
    \multirow{4}{*}{7} & \texttt{l2-last} & 3,189,454 & 18,387 & 750 & \multirow{4}{*}{7,999,809} & \multirow{4}{*}{40,685} & \multirow{4}{*}{1,425} \\
    & \texttt{l1-hidden\_a} & 3,858,217 & 20,179 & 823 & & & \\
    & \texttt{l1-hidden\_b} & 2,905,346 & 21,435 & 904 & & & \\
    & \texttt{l0-first} & 3,949,103 & 21,133 & 884 & & & \\
    \bottomrule
  \end{tabular}
\end{table}

\begin{table}[H]
  \centering
  \caption{WebShop optimizer cost by backward round.}
  \label{tab:optimizer-cost-webshop}
  \scriptsize
  \setlength{\tabcolsep}{1.9pt}
  \renewcommand{\arraystretch}{0.86}
  \begin{tabular}{@{}clrrrrrr@{}}
    \toprule
    & \multicolumn{4}{c}{\textbf{Hierarchical Optimizer}}
    & \multicolumn{3}{c}{\textbf{Global Optimizer}} \\
    \cmidrule(lr){2-5}\cmidrule(l){6-8}
    \textbf{Backward}
    & \textbf{Node}
    & \textbf{\shortstack{Input\\Token}}
    & \textbf{\shortstack{Output\\Token}}
    & \textbf{\shortstack{Optimizer\\Time (s)}}
    & \textbf{\shortstack{Input\\Token}}
    & \textbf{\shortstack{Output\\Token}}
    & \textbf{\shortstack{Optimizer\\Time (s)}} \\
    \midrule
    \multirow{4}{*}{1} & \texttt{l2-last} & 3,971,824 & 18,694 & 707 & \multirow{4}{*}{20,197,746} & \multirow{4}{*}{44,424} & \multirow{4}{*}{2,378} \\
    & \texttt{l1-hidden\_a} & 4,240,691 & 18,872 & 776 & & & \\
    & \texttt{l1-hidden\_b} & 2,977,722 & 14,114 & 603 & & & \\
    & \texttt{l0-first} & 5,586,749 & 22,601 & 870 & & & \\
    \cmidrule(lr){2-8}
    \multirow{4}{*}{2} & \texttt{l2-last} & 2,659,254 & 15,242 & 627 & \multirow{4}{*}{9,745,839} & \multirow{4}{*}{25,382} & \multirow{4}{*}{1,373} \\
    & \texttt{l1-hidden\_a} & 5,084,084 & 19,455 & 879 & & & \\
    & \texttt{l1-hidden\_b} & 3,130,255 & 17,547 & 766 & & & \\
    & \texttt{l0-first} & 3,556,342 & 16,677 & 723 & & & \\
    \cmidrule(lr){2-8}
    \multirow{4}{*}{3} & \texttt{l2-last} & 2,553,968 & 18,545 & 722 & \multirow{4}{*}{6,451,227} & \multirow{4}{*}{25,342} & \multirow{4}{*}{1,221} \\
    & \texttt{l1-hidden\_a} & 4,175,677 & 22,043 & 898 & & & \\
    & \texttt{l1-hidden\_b} & 4,877,493 & 24,138 & 1,011 & & & \\
    & \texttt{l0-first} & 3,651,828 & 20,312 & 812 & & & \\
    \cmidrule(lr){2-8}
    \multirow{4}{*}{4} & \texttt{l2-last} & 3,173,158 & 15,827 & 664 & \multirow{4}{*}{11,285,091} & \multirow{4}{*}{44,076} & \multirow{4}{*}{2,214} \\
    & \texttt{l1-hidden\_a} & 4,061,413 & 19,711 & 807 & & & \\
    & \texttt{l1-hidden\_b} & 4,403,027 & 19,539 & 841 & & & \\
    & \texttt{l0-first} & 4,278,530 & 16,154 & 697 & & & \\
    \cmidrule(lr){2-8}
    \multirow{4}{*}{5} & \texttt{l2-last} & 3,725,265 & 19,996 & 779 & \multirow{4}{*}{4,928,535} & \multirow{4}{*}{20,146} & \multirow{4}{*}{1,233} \\
    & \texttt{l1-hidden\_a} & 7,745,672 & 26,289 & 1,038 & & & \\
    & \texttt{l1-hidden\_b} & 4,451,916 & 22,171 & 858 & & & \\
    & \texttt{l0-first} & 5,969,030 & 22,942 & 1,011 & & & \\
    \cmidrule(lr){2-8}
    \multirow{4}{*}{6} & \texttt{l2-last} & 2,935,015 & 14,280 & 625 & \multirow{4}{*}{6,849,504} & \multirow{4}{*}{20,070} & \multirow{4}{*}{1,303} \\
    & \texttt{l1-hidden\_a} & 5,462,015 & 22,510 & 949 & & & \\
    & \texttt{l1-hidden\_b} & 3,445,449 & 15,389 & 728 & & & \\
    & \texttt{l0-first} & 3,231,752 & 16,726 & 722 & & & \\
    \cmidrule(lr){2-8}
    \multirow{4}{*}{7} & \texttt{l2-last} & 5,280,300 & 18,280 & 781 & \multirow{4}{*}{4,216,686} & \multirow{4}{*}{15,110} & \multirow{4}{*}{1,074} \\
    & \texttt{l1-hidden\_a} & 4,712,307 & 25,060 & 1,003 & & & \\
    & \texttt{l1-hidden\_b} & 4,858,977 & 19,708 & 846 & & & \\
    & \texttt{l0-first} & 3,708,947 & 17,573 & 771 & & & \\
    \bottomrule
  \end{tabular}
\end{table}

\clearpage
\subsection{Optimizer Prompt Templates}
\label{app:optimizer-ablation-prompts}

The templates below follow the latest semantic-report prompt wording in the
\texttt{Dev} implementation for each optimizer architecture.  To match the
comparison-study jobs, their execution controls are fixed to
\texttt{gpt-5.6-sol}, \texttt{reasoning\_effort=high}, and
\texttt{mode=standard}.  Apart from these study-specific control values, the
template text is reproduced verbatim; line wrapping and continuation indentation
are adjusted only for typesetting.  Per-job rendered prompts additionally append
the adopted replay/history protocol and the job-specific identifier and manifest
path.

\begin{lstlisting}[
  style=prompttemplate,
  caption={Hierarchical optimizer semantic-report prompt template.},
  label={lst:hierarchical-optimizer-prompt}
]
# Hierarchical node optimizer: semantic report

Execute only the frozen `gpt-5.6-sol`, `reasoning_effort=high`, `mode=standard` job using
  `delivery.protocol=semantic-report-v1`. Enforce `strict_local_visibility`: read the registered local failure
  bundles, direct downstream feedback, environment feedback when provided, checkpoint and fixed contract. The rendered
  prompt is read-only. Inspect `case_memory_snapshot` when diagnosing history or memory-continuity failures; do not
  inspect peer or non-neighbor private traces.

Before choosing a repair, systematically review the entire failure batch visible to this job. Check every case's
  feedback and key behaviors first, then inspect relevant traces and implementation more deeply by failure pattern.
  This does not require expanding every full model trace. Distinguish observed symptoms, root cause hypotheses and
  supporting evidence; do not force all failures into one explanation or stop the survey at the first easy fix.

Prioritize one or more patterns by their coverage, severity, strength of causal evidence, likely transfer to similar
  tasks and feasibility within the frozen budget. Frequency alone does not determine importance; no numerical scoring
  formula is required. Aim to improve the harness's general capability on similar tasks, beyond correcting isolated
  cases or passing selected replay samples. A single round need not resolve every pattern.

In the existing diagnosis, briefly summarize the main patterns, their supported scope and representative case IDs, why
  the selected targets deserve priority, and the validation findings and remaining uncertainties. Use existing tests
  and any adopted replay protocol to check the chosen solution. Distinguish observed local gains from expected broader
  benefits and note untested patterns and potential regressions. Do not create separate diagnostic reports or new
  report fields for this survey.

Distinguish four boundaries when choosing a repair:

- Fixed interfaces: preserve the fixed SDK, entrypoint, identity, request/response contract, authorized tool
  interfaces and execution-error propagation.
- Execution constraints: obey the frozen job's model and call budgets, visibility, read/write paths, validation and
  any adopted replay protocol.
- FH0 initialization limits: requirements for a neutral minimal baseline and initializer restrictions on template
  edits govern initialization. They do not freeze optimizer-trainable mechanisms after initialization. Continue to
  honor the fixed contract; existing README/ROLE descriptions of baseline behavior do not independently make an
  allowed implementation path immutable.
- Trainable decision mechanisms: repository `allowed_paths` are trainable paths. Changes to role
  definitions/descriptions and prompts, as well as code-level changes to strategy, memory, parsing or tools, are
  permitted where declared. Choose the mechanism from the observed failure evidence; the FH0 role split is not
  immutable, but fixed topology, interfaces and tool permissions remain binding.

When needed to address the prioritized failure patterns, you may change or replace existing mechanisms, restructure
  trainable code, or introduce a new mechanism or design within those paths and boundaries. Existing code need not
  have an implementation bug to justify a code-level decision-policy improvement. Prefer the smallest change necessary
  to address the failure mechanism, not the fewest edited files or a prompt-only fix. Do not require a prompt attempt
  before considering code, or change code merely to increase the number of code edits.

Keep the optimized harness's forward input context and node outputs lightweight while preserving decision-critical
  information, honoring the fixed contract and achieving the intended mechanism improvement. Avoid redundant
  information injection, unnecessary expansion of history or evidence, and accumulating verbose rules or intermediate
  explanations. Pass only the information the downstream consumer needs to fulfill its role. Consider both input and
  output costs, including downstream reuse and accumulation in later history. Additional context or output should have
  a clear mechanism benefit. This is an optimization preference, not a hard token limit or a new acceptance criterion.
  Never discard key constraints, evidence, uncertainty or required history merely to reduce tokens; continue to obey
  the frozen history contract.

Apply one candidate update to the published base, run necessary development tests, and leave the candidate repository
  clean. Use the interpreter provided by the controller, with the candidate repository as the test cwd. For
  `no_change`, leave the published base unchanged. Keep the fixed contract and all model and visibility limits.

Keep test source files in their registered allowed paths. Put test-generated temporary files under the job-specific
  `Dev/tmp/` directory given in the launch instruction. When running pytest from the candidate repository, pass that
  directory as `--basetemp`; do not create test temporary directories inside the candidate repository. The runner owns
  cleanup of its job-specific directory.

The semantic report is the only final deliverable; this does not prohibit the authorized candidate edits,
  development tests, or adopted replay input files and controller replay commands. Complete those steps before writing
  one JSON report to `delivery.report_path`, following `delivery.report_schema`. Successful reports contain
  `schema_version="1.0"`, `job_id`, `status="succeeded"`, `diagnosis`, `evidence_case_ids`,
  `consumed_environment_feedback_ids`, one `agent_decisions` entry, `consumed_downstream_feedback_ids`, and
  `upstream_feedback`. The decision names `agent_id`, `outcome` (`changed` or `no_change`), `change_summary`, and its
  own `evidence_case_ids`. Evidence case lists must be nonempty, unique subsets of the visible cases. Confirm
  consumption of every declared environment and downstream feedback; use environment case IDs and downstream feedback
  IDs. Keep diagnosis and summaries brief and evidence-based.

Provide exactly one upstream feedback entry per direct recipient, even for `no_change`. Each entry has
  `recipient_agent_id`, `feedback_type`, `case_ids`, `summary`, `improvement_points`, and `preserve_points`. `improve`
  requires improvement points; `preserve` requires preserve points and no improvement points; `no_special_guidance`
  has neither. Explain useful upstream input without single-point blame or claims of single-node causality for an
  end-to-end failure.

If unable to finish, write only `schema_version`, `job_id`, `status="failed"`, and a nonempty `error`. After the
  candidate, necessary tests and report are complete, finish immediately. The controller owns declared delivery
  checks, diff capture, inventory, execution statistics and full result generation. Do not generate those
  artifacts, write a schema checker, invoke acceptance, activate a checkpoint, or retry the job.
\end{lstlisting}

\begin{lstlisting}[
  style=prompttemplate,
  caption={Global (flat-network) optimizer semantic-report prompt template.},
  label={lst:flat-optimizer-prompt}
]
# Flat network optimizer: semantic report

Execute only the frozen `gpt-5.6-sol`, `reasoning_effort=high`, `mode=standard` job using
  `delivery.protocol=semantic-report-v1`. Read the registered whole-network failure bundles, environment feedback,
  checkpoint and fixed contract. The rendered prompt is read-only. Inspect `case_memory_snapshot` when diagnosing
  history or memory-continuity failures; use only this job's visible evidence.

Before choosing a repair, systematically review the entire failure batch visible to this job. Check every case's
  feedback and key behaviors first, then inspect relevant traces and implementation more deeply by failure pattern.
  This does not require expanding every full model trace. Distinguish observed symptoms, root cause hypotheses and
  supporting evidence; do not force all failures into one explanation or stop the survey at the first easy fix.

Prioritize one or more patterns by their coverage, severity, strength of causal evidence, likely transfer to similar
  tasks and feasibility within the frozen budget. Frequency alone does not determine importance; no numerical scoring
  formula is required. Aim to improve the harness's general capability on similar tasks, beyond correcting isolated
  cases or passing selected replay samples. A single round need not resolve every pattern.

In the existing diagnosis, briefly summarize the main patterns, their supported scope and representative case IDs, why
  the selected targets deserve priority, and the validation findings and remaining uncertainties. Use existing tests
  and any adopted replay protocol to check the chosen solution. Distinguish observed local gains from expected broader
  benefits and note untested patterns and potential regressions. Do not create separate diagnostic reports or new
  report fields for this survey.

Distinguish four boundaries when choosing a repair:

- Fixed interfaces: preserve the fixed SDK, entrypoint, identity, request/response contract, authorized tool
  interfaces and execution-error propagation.
- Execution constraints: obey the frozen job's model and call budgets, visibility, read/write paths, validation and
  any adopted replay protocol.
- FH0 initialization limits: requirements for a neutral minimal baseline and initializer restrictions on template
  edits govern initialization. They do not freeze optimizer-trainable mechanisms after initialization. Continue to
  honor the fixed contract; existing README/ROLE descriptions of baseline behavior do not independently make an
  allowed implementation path immutable.
- Trainable decision mechanisms: repository `allowed_paths` are trainable paths. Changes to role
  definitions/descriptions and prompts, as well as code-level changes to strategy, memory, parsing or tools, are
  permitted where declared. Choose the mechanism from the observed failure evidence; the FH0 role split is not
  immutable, but fixed topology, interfaces and tool permissions remain binding.

When needed to address the prioritized failure patterns, you may change or replace existing mechanisms, restructure
  trainable code, or introduce a new mechanism or design within those paths and boundaries. Existing code need not
  have an implementation bug to justify a code-level decision-policy improvement. Prefer the smallest change necessary
  to address the failure mechanism, not the fewest edited files or a prompt-only fix. Do not require a prompt attempt
  before considering code, or change code merely to increase the number of code edits.

Keep the optimized harness's forward input context and node outputs lightweight while preserving decision-critical
  information, honoring the fixed contract and achieving the intended mechanism improvement. Avoid redundant
  information injection, unnecessary expansion of history or evidence, and accumulating verbose rules or intermediate
  explanations. Pass only the information the downstream consumer needs to fulfill its role. Consider both input and
  output costs, including downstream reuse and accumulation in later history. Additional context or output should have
  a clear mechanism benefit. This is an optimization preference, not a hard token limit or a new acceptance criterion.
  Never discard key constraints, evidence, uncertainty or required history merely to reduce tokens; continue to obey
  the frozen history contract.

Apply one candidate update per changed node to its published base, run necessary development tests, and leave all
  candidate repositories clean. Use the interpreter provided by the controller, with the candidate repository as the
  test cwd. For `no_change`, leave the published base unchanged. Keep the fixed contract and all model limits.

Keep test source files in their registered allowed paths. Put test-generated temporary files under the job-specific
  `Dev/tmp/` directory given in the launch instruction. When running pytest from a candidate repository, pass that
  directory as `--basetemp`; do not create test temporary directories inside a candidate repository. The runner owns
  cleanup of its job-specific directory.

The semantic report is the only final deliverable; this does not prohibit the authorized candidate edits,
  development tests, or adopted replay input files and controller replay commands. Complete those steps before writing
  one JSON report to `delivery.report_path`, following `delivery.report_schema`. Successful reports contain
  `schema_version="1.0"`, `job_id`, `status="succeeded"`, `diagnosis`, `evidence_case_ids`,
  `consumed_environment_feedback_ids`, and `agent_decisions`. Include exactly one decision per target node, naming
  `agent_id`, `outcome` (`changed` or `no_change`), `change_summary`, and that node's own `evidence_case_ids`.
  Evidence case lists must be nonempty, unique subsets of visible cases. Confirm consumption of every declared
  environment feedback using its case ID. Keep diagnosis and summaries brief and evidence-based. Flat jobs must omit
  upstream feedback and consumed downstream feedback fields; the controller derives the network outcome and candidate
  checkpoint from decisions and accepted candidate changes.

If unable to finish, write only `schema_version`, `job_id`, `status="failed"`, and a nonempty `error`. After the
  candidates, necessary tests and report are complete, finish immediately. The controller owns declared delivery
  checks, diff capture, inventory, execution statistics and full result generation. Do not generate those
  artifacts, write a schema checker, invoke acceptance, activate a checkpoint, or retry the job.
\end{lstlisting}

\endgroup

\clearpage
\section{Case Study: Learn Rules, Not Answers}
\label{app:case-study}

We examine whether the performance gains come from reusable task-specific
mechanisms encoded in the evolved harnesses rather than memorized answers to
training tasks. We analyze the 4-node ALFWorld experiment with Qwen3.5-9B,
shown in the corresponding training-results figure in the main paper. The
foundation models remain frozen, and all updates are confined to the forward
harness repositories. This makes the evolution process directly inspectable.
We divide the process into four stages based on the training and validation
curves; stage headings report checkpoint validation pass rates.

\textbf{FH0 Initialization.}
The initial HarnessNet uses a 4-node topology with the following role division:
\begin{itemize}
\item \textbf{Input Node:} Organizes the task, current observation, interaction history, and admissible actions for downstream nodes.
\item \textbf{Hidden Node A:} Independently analyzes which actions are currently feasible and checks their preconditions.
\item \textbf{Hidden Node B:} Tracks goal progress and recommends the next useful transition toward task completion.
\item \textbf{Output Node:} Reconciles the two hidden-node analyses and selects one action from the current admissible set.
\end{itemize}
At FH0, these nodes contain only general role prompts and forwarding logic. They
do not have explicit mechanisms for loop prevention, object-state tracking,
multi-object progress, or search planning.

\textbf{FH0 (19.05\%) $\rightarrow$ FH1 (33.33\%).}
Training and validation performance improve sharply in this stage. Unreliable
environment interaction is the main FH0 failure mode: 25 of the first 40 failed
cases terminate with invalid actions. Updates in FH1 include:
\begin{itemize}
\item \textbf{Input Node:} Preserves the latest observation and records held objects and searched locations.
\item \textbf{Hidden Node A:} Detects repeated information-gathering actions to avoid wasting steps.
\item \textbf{Hidden Node B:} Recommends canonical actions, begins tracking object counts and transformation states, and enforces the one-object inventory constraint.
\item \textbf{Output Node:} Converts malformed suggestions into canonical commands, rejects unavailable actions, and resolves conflicts using current environment evidence.
\end{itemize}
These changes establish more reliable multi-step interaction with the
environment. Invalid-action failures decrease from 25 to 5 in the next failure
batch, shifting the main bottleneck from action validity to long-horizon task
completion.

\textbf{FH1 (33.33\%) $\rightarrow$ FH4 (38.10\%).}
Performance improves gradually and non-monotonically. FH1 often searches while
already holding the target, confuses similar object names or action meanings,
and allows old history to override the latest location or inventory. Updates
through FH4 include:
\begin{itemize}
\item \textbf{Input Node:} Introduces a Goal$\rightarrow$Now$\rightarrow$Progress$\rightarrow$Next$\rightarrow$Later structure to organize state and action order.
\item \textbf{Hidden Node A:} Uses the latest location and inventory as state and checks immediate action preconditions.
\item \textbf{Hidden Node B:} Preserves completed deliveries, tracks held and distinct object instances, and requires observed evidence before marking transformations complete.
\item \textbf{Output Node:} Uses the current state and action history to avoid cycles, unnecessary searches, and redundant actions.
\end{itemize}
By FH4, all 40 failures reach the step limit rather than terminating with invalid
actions. Ten cases make partial progress on pickup, transformation, or placement,
while the other 30 perform valid but overly long searches. Thus, training
reduces state and action errors but exposes search efficiency as the next
bottleneck.

\textbf{FH4 (38.10\%) $\rightarrow$ FH5 (76.19\%).}
This stage yields the largest training and validation gains. All 40 FH4 failures
reach the step limit, and 16 cases repeatedly search locations of the same kind
instead of trying another plausible source. Updates in FH5 include:
\begin{itemize}
\item \textbf{Input Node:} Learns source-family priors for common target objects and switches families after repeated misses.
\item \textbf{Hidden Node A:} Detects search stagnation and proposes an unvisited location from a different plausible family.
\item \textbf{Hidden Node B:} Refreshes its plan from the latest visible objects and visited locations, prioritizing visible targets and avoiding unsupported locations.
\item \textbf{Output Node:} Remains unchanged.
\end{itemize}
In the next 35-case failure batch, no trajectory visits three or more distinct
locations from one family consecutively before switching families. This greatly
improves training and validation performance.

\textbf{FH5 (76.19\%) $\rightarrow$ FH7 (83.33\%).}
Performance improves from FH5 to FH6 and then saturates, with no further
validation gain at FH7. Remaining failures include leaving plausible source
locations too early, attempting remote examination, repeating stale
destinations, treating proposed actions as completed, and selecting the same
object instance again. Updates through FH7 include:
\begin{itemize}
\item \textbf{Input Node:} Prioritizes unvisited sources and stops repeating ignored destinations.
\item \textbf{Hidden Node A:} Distinguishes proposed actions from executed actions and analyzes action feasibility.
\item \textbf{Hidden Node B:} Tracks completed deliveries and individual object instances, avoids stale search locations, and includes required steps such as opening an appliance before using it.
\item \textbf{Output Node:} Uses Node A's feasibility analysis to correct Node B's plan when a prerequisite is missing.
\end{itemize}
These changes refine search ordering and multi-step execution.  They produce the
smaller improvement from FH5 to FH6, while FH7 maintains the same validation
pass rate.

The evolution process preserves the FH0 node roles while adding
complementary ALFWorld-specific mechanisms inside each node forward harness.
The final harnesses contain neither case IDs nor stored case-specific action
sequences. The best evolved checkpoint also improves held-out test
performance from 16.51\% at FH0 to 69.72\% at FH6.  Together, the absence of
stored case sequences and the held-out improvement suggest that HarnessNet
learns reusable interaction rules rather than memorizing training answers.

\end{document}